\documentclass{article} 
\usepackage{iclr2019_conference,times}

\usepackage{amsmath,amsfonts,bm}

\def\eqref#1{equation~\ref{#1}}

\def\1{\bm{1}}

\DeclareMathAlphabet{\mathsfit}{\encodingdefault}{\sfdefault}{m}{sl}
\SetMathAlphabet{\mathsfit}{bold}{\encodingdefault}{\sfdefault}{bx}{n}

\DeclareMathOperator{\sign}{sign}

\newcommand{\epspack}[4]{
\begin{figure}[htbp]
    \centering
    \begin{subfigure}[b]{0.49\textwidth}
        \includegraphics[width=\textwidth]{charts/#1/#2/logit.pdf}
    \end{subfigure}
    \begin{subfigure}[b]{0.49\textwidth}
        \includegraphics[width=\textwidth]{charts/#1/#2/eps.pdf}
    \end{subfigure}
        \centering
    \begin{subfigure}[b]{0.49\textwidth}
        \includegraphics[width=\textwidth]{charts/#1/#2/grad.pdf}
    \end{subfigure}
    \begin{subfigure}[b]{0.49\textwidth}
        \includegraphics[width=\textwidth]{charts/#1/#2/cosine.pdf}
    \end{subfigure}
    \caption{#3}
    \label{#4}
\end{figure}
}

\usepackage[hidelinks]{hyperref}
\usepackage{url}

\usepackage{comment}
\usepackage{graphics}
\usepackage{graphicx}
\usepackage{subcaption}
\usepackage[noabbrev,capitalize]{cleveref} 

\usepackage{multirow}
\usepackage{multicol}

\usepackage{enumitem}

\newcommand{\bx}{\bm{x}}
\newcommand{\by}{\bm{y}}
\newcommand{\bz}{\bm{z}}
\newcommand{\btheta}{\bm{\theta}}
\newcommand{\bdelta}{\bm{\delta}}

\newcommand{\mytitle}{Label Smoothing and Logit Squeezing: A Replacement for Adversarial Training?}

\title{\mytitle}

\author{Ali Shafahi, Amin Ghiasi, Furong Huang \& Tom Goldstein \\
Department of Computer Science\\
University of Maryland \\
College Park, Maryland, USA \\
\texttt{\{ashafahi,amin,furongh,tomg\}@cs.umd.edu} \\
}

\iclrfinalcopy 
\begin{document}

\maketitle

\begin{abstract}
 Adversarial training is one of the strongest defenses against adversarial attacks, but it requires adversarial examples to be generated for every mini-batch during optimization.  The expense of producing these examples during training often precludes adversarial training from use on complex image datasets. 
 In this study, we explore the mechanisms by which adversarial training improves classifier robustness, and show that these mechanisms can be effectively mimicked using simple regularization methods, including label smoothing and logit squeezing.  
 Remarkably, using these simple regularization methods in combination with Gaussian noise injection, we are able to achieve strong adversarial robustness -- often exceeding that of adversarial training -- using no adversarial examples. 
\end{abstract}

\section{Introduction}
Deep Neural Networks (DNNs) have enjoyed great success in many areas of computer vision,
 such as classification \citep{krizhevsky2012imagenet}, object detection \citep{girshick2015fast}, and face recognition \citep{najibi2017ssh}.
However, the existence of adversarial examples has raised concerns about the security of computer vision systems  \citep{szegedy2013intriguing,biggio2013evasion}. For example, an attacker may cause a system to mistake a stop sign for another object \citep{evtimov2017robust} or mistake one person for another \citep{sharif2016accessorize}. To address security concerns for high-stakes applications, researchers are searching for ways to make models more robust to attacks.

Many defenses have been proposed to combat adversarial examples.   Approaches such as feature squeezing, denoising, and encoding \citep{xu2017feature, samangouei2018defense, shen2017ape, meng2017magnet} have had some success at pre-processing images to remove adversarial perturbations. Other approaches focus on hardening neural classifiers to reduce adversarial susceptibility.  This includes specialized non-linearities \citep{zantedeschi2017efficient}, modified training processes \cite{papernot2016distillation}, and gradient obfuscation \cite{DBLP:journals/corr/abs-1802-00420}. 

Despite all of these innovations, adversarial training \citep{goodfellow2014explaining}, one of the earliest defenses, still remains among the most effective and popular strategies. 
In its simplest form, adversarial training minimizes a loss function that measures performance of the model on both clean and adversarial data as follows

\begin{equation}
    \underset{\theta}{\mathrm{minimize}} 
    \hspace{1.5mm}
    \textbf{$L_{adv}(\btheta)$} = \sum_i  \kappa \textbf{$L(\btheta, \bx_i, \by_i)$} + (1 - \kappa) \textbf{$L(\btheta, \bx_{i,adv}, \by_i)$},
\label{eq:adv_training}
\end{equation}
where $L$ is a standard (cross entropy) loss function, ($\bx_i$, $\by_i$) is an input image/label pair, $\theta$ contains the classifier's trainable parameters, $\kappa$ is a hyper-parameter, and $\bx_{i,adv}$ is an adversarial example for image $\bx$. 
\cite{madry2017towards} pose adversarial training as a game between two players that similarly requires computing adversarial examples on each iteration.

A key drawback to adversarial training methods is their computational cost; after every mini-batch of training data is formed, a batch of adversarial examples must be produced. 
To train a network that resists strong attacks, one needs to train with the strongest adversarial examples possible. For example, networks hardened against the inexpensive Fast Gradient Sign Method (FGSM, \cite{goodfellow2014explaining}) can be broken by a simple two-stage attack \citep{tramer2017ensemble}. 
Current state-of-the-art adversarial training results on  MNIST and CIFAR-10 use expensive iterative adversaries \citep{madry2017towards}, such as the Projected Gradient Descent (PGD) method, or the closely related Basic Iterative Method (BIM) \citep{kurakin2016adversarialBIM}. Adversarial training using strong attacks may be 10-100 times more time consuming than standard training methods.
This prohibitive cost makes it difficult to scale adversarial training to larger datasets and higher resolutions.

In this study, we show that it is possible to achieve strong robustness -- comparable to or greater than the robustness of adversarial training with a strong iterative attack -- using fast optimization without adversarial examples.  We achieve this using standard regularization methods, such as label smoothing \citep{warde201611} and the more recently proposed logit squeezing \citep{kannan2018adversarial}.  While it has been known for some time that these tricks can improve the robustness of models, we observe that an aggressive application of these inexpensive tricks, combined with random Gaussian noise, are enough to match or even surpass the performance of adversarial training on some datasets.  For example, using only label smoothing and augmentation with random Gaussian noise, we produce a CIFAR-10 classifier that achieves over 73\% accuracy against black-box iterative attacks, compared to 64\% for a state-of-the-art adversarially trained classifier \citep{madry2017towards}. In the white-box case, classifiers trained with logit squeezing and label smoothing get $\approx$ 50\% accuracy on iterative attacks in comparison to $\approx$ 47\% for adversarial training. Regularized networks without adversarial training are also more robust against non-iterative attacks, and more accurate on non-adversarial examples.

Our goal is not just to demonstrate these defenses, but also to dig deep into what adversarial training does, and how it compares to less expensive regularization-based defenses.  We begin by dissecting adversarial training, and examining ways in which it achieves robustness. We then discuss label smoothing and logit squeezing regularizers, and how their effects compare to those of adversarial training. We then turn our attention to random Gaussian data augmentation, and explore the importance of this technique for adversarial robustness.  Finally, we combine the regularization methods with random Gaussian augmentation, and experimentally compare the robustness achievable using these simple methods to that achievable using adversarial training.

\section{What does adversarial training do?}

Adversarial training injects adversarial examples into the training data as SGD runs.  During training, adversarial perturbations are applied to each training image to decrease the logit corresponding to its correct class.  The network must learn to produce logit representations that preserve the correct labeling even when faced with such an attack.
At first glance, it seems that adversarial training might work by producing a large ``logit gap,'' i.e., by producing a logit for the true class that is much larger than the logit of other classes.  Surprisingly, adversarial training has the opposite effect -- we will see below that it {\em decreases} the logit gap.
To better understand what adversarial training does, and how we can replicate it, we now break down the different strategies for achieving robustness.

\subsection{A simple linearized model of adversarial robustness}

This section presents a simple metric for adversarial robustness that will help us understand adversarial training.  
Consider an image $\bx$, and its logit representation $\bz$ (i.e. pre-softmax activation) produced by a neural network.   Let $z_y$ denote the logit corresponding to the correct class $y.$ If we add a small perturbation $\bdelta$ to $\bx,$ then the corresponding change in logits is approximately $\bdelta^T \nabla_x z_y$ under a linear approximation, where $\nabla_x z_y$ is the gradient of $z_y$ with respect to $\bx.$

Under a linearity assumption, we can calculate the step-size $\epsilon_L$ needed to move an example from class $y$ to another class $\Bar{y}.$ 
A classifier is susceptible to adversarial perturbation $\bdelta$ if the perturbed logit of the true class is smaller than the perturbed logit of any other class:
\begin{equation}
    z_y + \bdelta^ T \nabla_x z_y < z_{\Bar{y}} + \bdelta^ T \nabla_x z_{\Bar{y}}.
    \label{eq:equal_logits}
\end{equation}
Assuming a one-step $\ell_\infty$ attack such as FGSM, the perturbation $\bdelta$ can be expressed as
\begin{equation}
    \bdelta = -\epsilon_L \cdot \sign (\nabla_x z_y - \nabla_x z_{\Bar{y}} ),
    \label{eq:delta_wrt_logit}
\end{equation}
where $\epsilon_L$ is the $\ell_\infty$-norm of the perturbation. 
Using this choice of $\bdelta,$  Equation \ref{eq:equal_logits} becomes
\begin{align}
    z_y - z_{\Bar{y}} &< -\epsilon_L \cdot \sign(\nabla_x z_y - \nabla_x z_{\Bar{y}})^T (\nabla_x z_{\Bar{y}} - \nabla_x z_y) \nonumber  = \epsilon_L \|\nabla_x z_y - \nabla_x z_{\Bar{y}} \|_1 
\end{align} 
where $\|\cdot\|_1$ denotes the $\ell_1$-norm of a vector. Therefore the smallest $\ell_\infty$-norm of the perturbation required is the ratio of ``logit gap'' to ``gradient gap'', i.e.,
\begin{align} 
    \epsilon_L &>  \frac{z_y - z_{\Bar{y}}}{\|\nabla_x z_y - \nabla_x z_{\Bar{y}}\|_1}.
    \label{eq:eps_smallest}
\end{align}

Equation \ref{eq:eps_smallest} measures robustness by predicting the smallest perturbation size needed to switch the class of an image. While the formula for $\epsilon_L$ makes linearity assumptions, the approximation $\epsilon_L$ fairly accurately predicts the robustness of classifiers of the CIFAR-10 dataset (where perturbations are small and linearity assumptions cause little distortion).  It is also a good ballpark approximation on MNIST, even after adversarial training (see Table \ref{tab:exp_analyt_eps}).

\begin{table}
    \centering
    \begin{tabular}{c|c|c|c}
    \multirow{2}{*}{description} &  \multicolumn{2}{c}{empirical FGSM} &  \multirow{2}{*}{Equation~\ref{eq:eps_smallest}} \\
    & on X-ent &  on logits (CW) & \\
    \hline  
    MNIST naturally trained ($\epsilon=0.3$) & 7.05\% & 0.23\% & 0.0\% \\
    MNIST adv trained ($\epsilon=0.3$) & 95.25\% & 95.41\% & 56.73\% \\
    CIFAR10 naturally trained ($\epsilon=8(/255)$) & 13.33\% & 10.64\% & 0.0\% \\
    CIFAR10 adv trained ($\epsilon=8(/255)$) & 56.22\% & 55.57\% & 54.97\% \\
    \end{tabular}
    \caption{Experimental and predicted accuracy of classifiers for MNIST and CIFAR-10. The predicted accuracy is the percentage of images for which $\epsilon<\epsilon_L$. The empirical accuracy is the percent of images that survive a perturbation of size $\epsilon$. Attacks on both the cross-entropy (X-ent) and logits as in \cite{carlini2017towards} (CW) are presented.}
    \label{tab:exp_analyt_eps}
\end{table}

Maximal robustness occurs when $\epsilon_L$ is as large as possible.  From \eqref{eq:eps_smallest}, we observe 3 different strategies for hardening a classifier:
\begin{itemize}[leftmargin=3mm]
    \item \textbf{Increase the logit gap}:  Maximize the numerator of equation \ref{eq:eps_smallest} by producing a classifier with relatively large $z_y.$  
    \item \textbf{Squash the adversarial gradients}:  Train a classifier that has small adversarial gradients $\nabla_x z_{\Bar{y}}$ for any class $\bar{y}.$  In this case a large perturbation is needed to significantly change $z_{\Bar{y}}.$ 
    \item \textbf{Maximize gradient coherence}: Produce adversarial gradients $\nabla_x z_{\Bar{y}}$ that are highly correlated with the gradient for the correct class $\nabla_x z_y.$ This will shrink the denominator of \eqref{eq:eps_smallest}, and produce robustness even if adversarial gradients are large. In this case, one cannot decrease $z_y$ without also decreasing $z_{\Bar{y}},$ and so large perturbations are needed to change the class label.
\end{itemize}

\begin{figure}[!h]
    \centering
    \begin{subfigure}[b]{0.49\textwidth}
        \includegraphics[width=\textwidth]{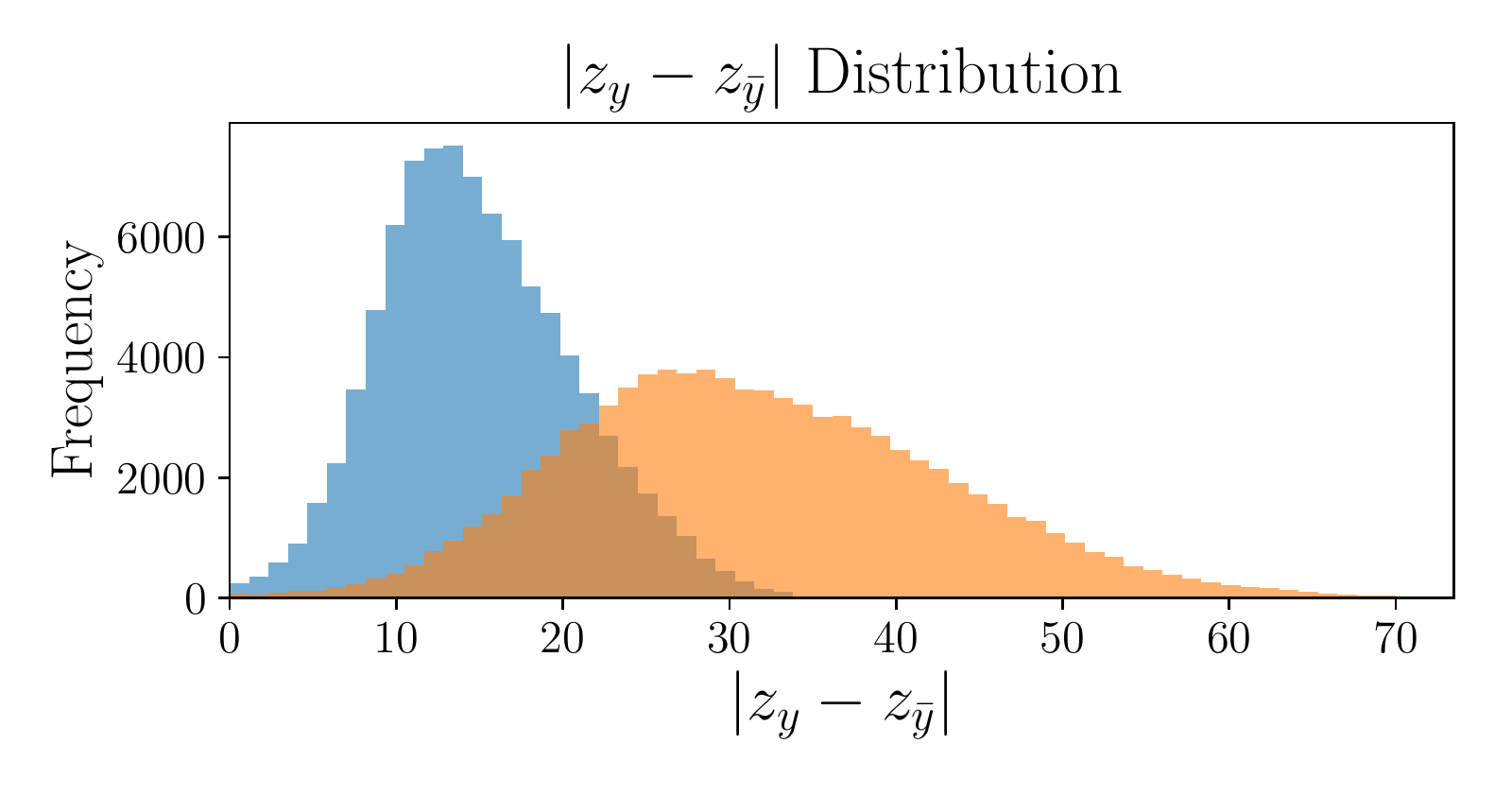}
    \end{subfigure}
    \begin{subfigure}[b]{0.49\textwidth}
        \includegraphics[width=\textwidth]{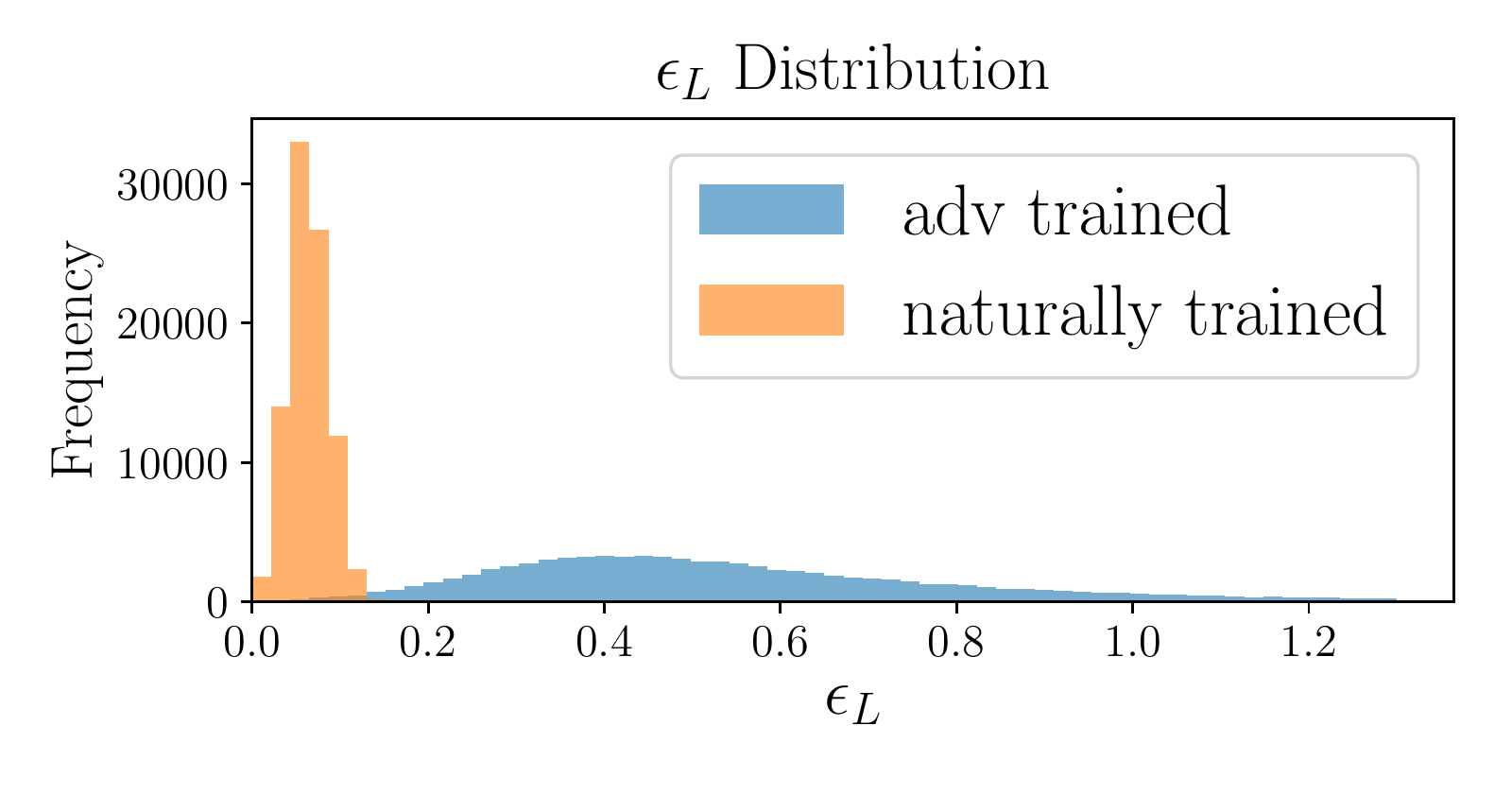}
    \end{subfigure}
        \centering
    \begin{subfigure}[b]{0.49\textwidth}
        \includegraphics[width=\textwidth]{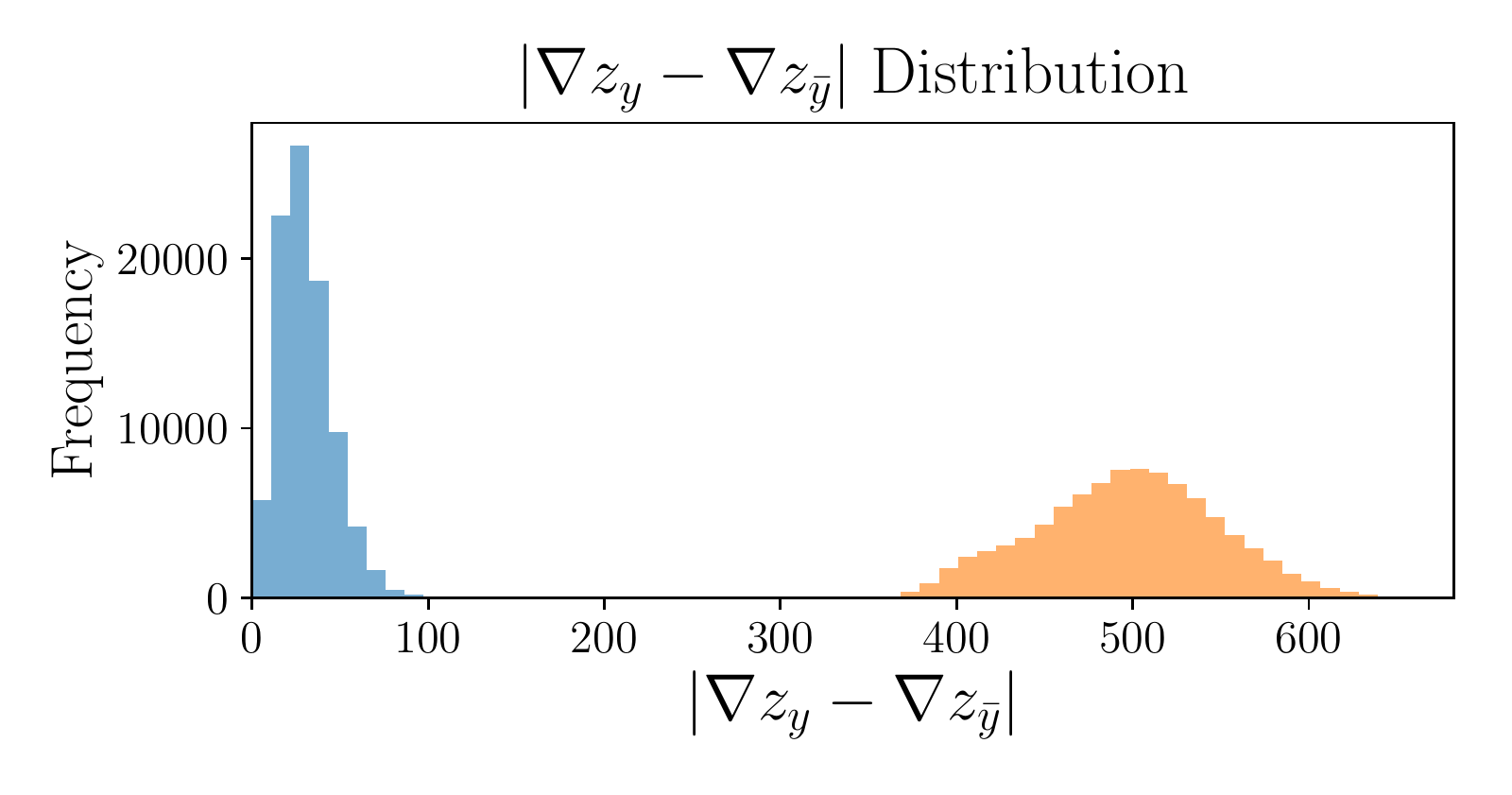}
    \end{subfigure}
    \begin{subfigure}[b]{0.49\textwidth}
        \includegraphics[width=\textwidth]{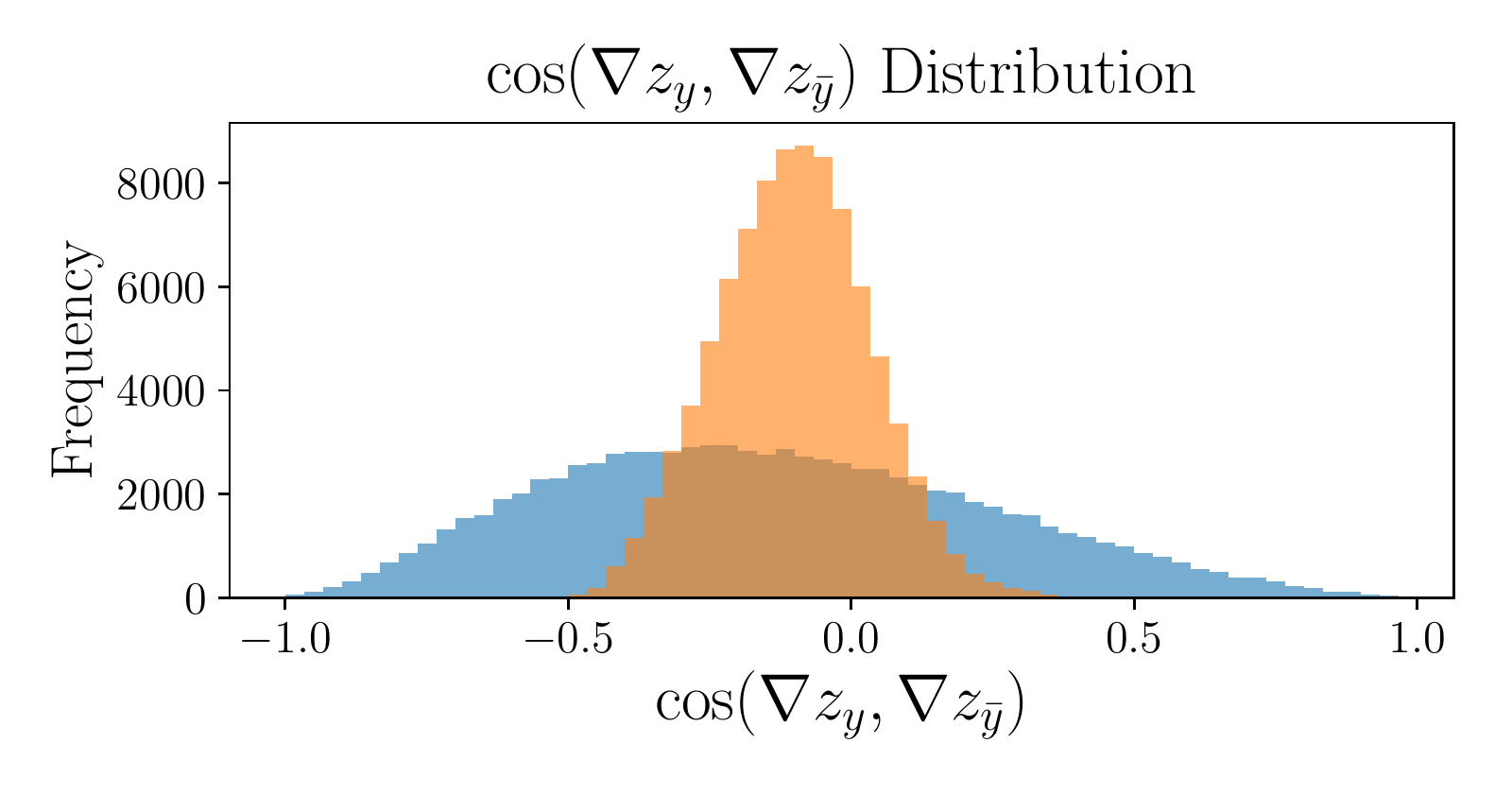}
    \end{subfigure}
    \caption{The effect of adversarial training on the numerator and denominator for $\epsilon_L$ on MNIST. (left-top) The numerator of Equation~\ref{eq:eps_smallest} (i.e, the logit gap). (left-bottom) The denominator of Equation~\ref{eq:eps_smallest}. (right-top) A histogram of values of $\epsilon_L$ calculated by applying Equation~\ref{eq:eps_smallest} to test data. (right-bottom) Cosine between the gradient vectors of the logits (i.e., gradient coherence).}
    \label{fig:mnist_eps_natNadv}
\end{figure}

The most obvious strategy for achieving robustness is to increase the numerator in equation \ref{eq:eps_smallest} while fixing the denominator.
Remarkably, our experimental investigation reveals that adversarial training does not rely on this strategy at all, but rather it decreases the logit gap and gradient gap simultaneously.
This can be observed in Figure \ref{fig:mnist_eps_natNadv}, which shows distributions of logit gaps for naturally and adversarially trained models on MNIST.  Note that the cross entropy loss actually limits adversarial training from increasing logit gaps.  The accuracy of the classifier goes down in the presence of adversarial examples, and so the cross entropy loss is minimized by smaller logit gaps that reflect the lower level of certainty in the adversarial training environment.

\begin{figure}[!htbp]
    \centering
    \begin{subfigure}[b]{0.49\textwidth}
        \includegraphics[width=\textwidth]{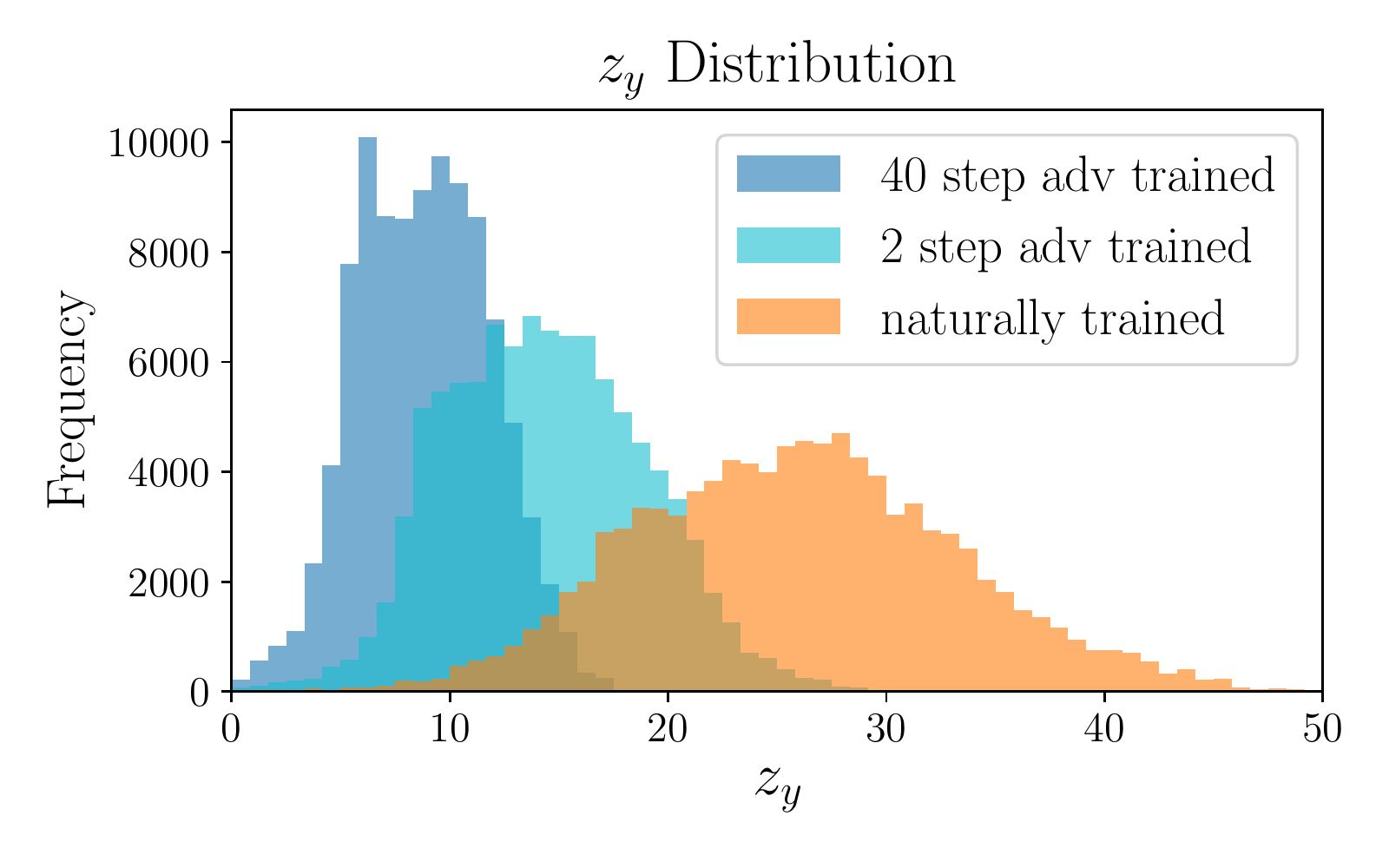}
        \caption{MNIST logit distributions.}
        \label{fig:mnist_logit_dist}
    \end{subfigure}
    \begin{subfigure}[b]{0.49\textwidth}
        \includegraphics[width=\textwidth]{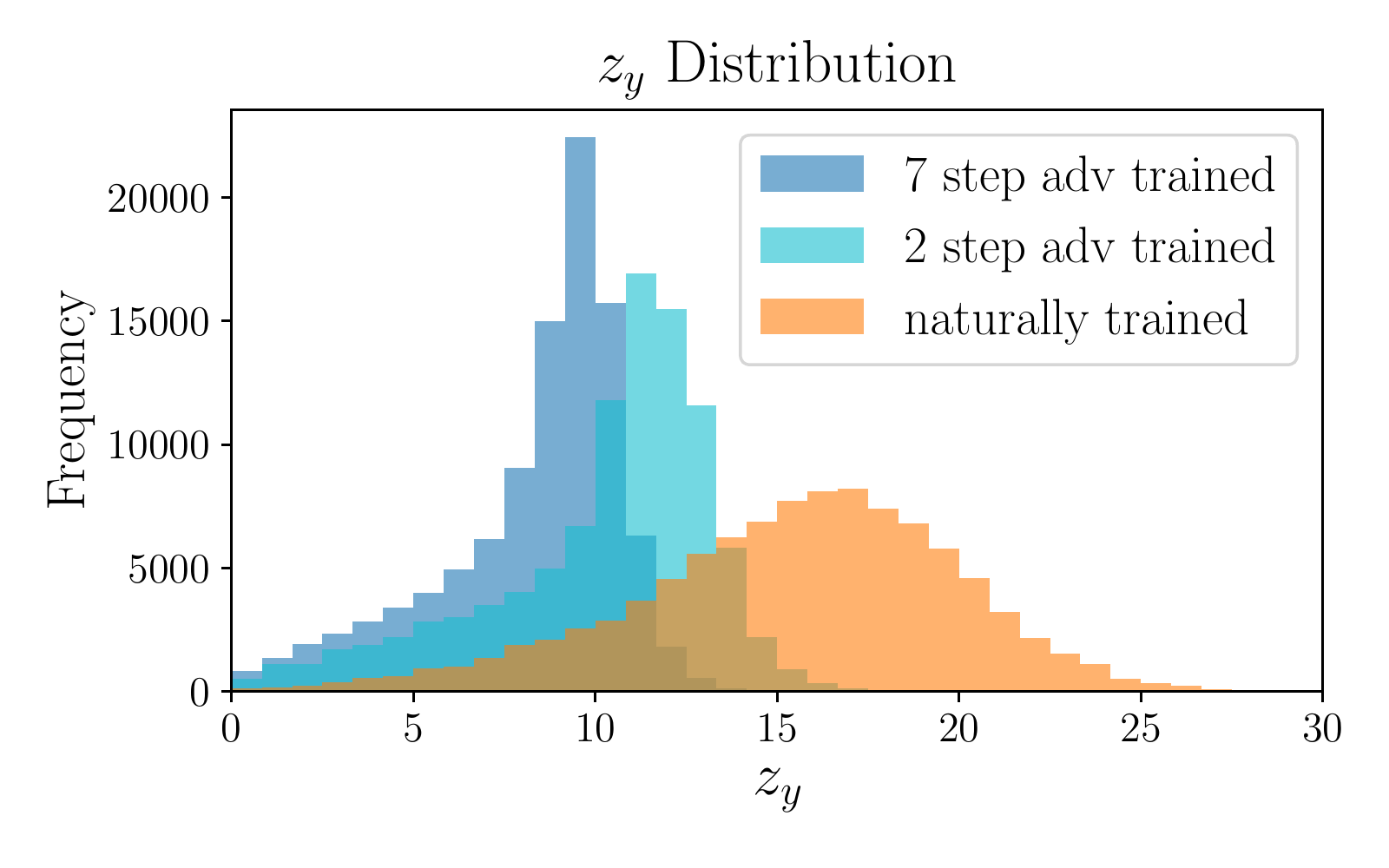}
        \caption{CIFAR10 logit distributions.}
        \label{fig:c10_logit_dist}
    \end{subfigure}
    \caption{Adversarial training squishes the logits. Training on stronger adversaries squishes more.}
    \label{fig:logit_dist_c10Nmnist}
\end{figure}

Adversarial training succeeds by minimizing the denominator in Equation \ref{eq:eps_smallest}; it simultaneously squeezes the logits and crushes the adversarial gradients.  Figure \ref{fig:mnist_eps_natNadv} shows that the adversarial gradients shrink dramatically more than the logit gaps, and so the net effect is an increase in robustness.

If we closely examine the phenomenon of shrinking the logit gaps, we find that this shrink is due in part to an overall shrink in the size of the logits themselves, (i.e., $|z_i|$ for any class $i$).
To see this, we plot histograms of the logits when classifiers are adversarially trained with strong adversaries\footnote{MNIST: 40-step PGD with step-size 0.01 and $\epsilon$=0.3. CIFAR-10: 7-step PGD with step-size 2 and $\epsilon=8$.}, weak adversaries\footnote{MNIST: 2-step PGD with step-size 0.2 and $\epsilon$=0.3. CIFAR-10: 2-step PGD with step-size 5 and $\epsilon=8$.}, and with no adversarial examples.  Figure \ref{fig:logit_dist_c10Nmnist} shows that adversarial training does indeed squash the logits, although not enough to fully explain the decrease in $| z_y - z_{\Bar{y}}|$  in Figure \ref{fig:mnist_eps_natNadv}.~\footnote{Similar plots for the effect of adversarial training on CIFAR-10 are in \cref{appendix:cifar10_natNadv} (\cref{fig:c10_eps_natNadv}).}

We have seen that adversarial training works by squashing adversarial gradients and slightly increasing gradient coherence. But the fact that it cannot do this without decreasing the logit gap leads us to suspect that these quantities are inherently linked.  This leads us to ask an important question:  If we directly decrease the logit gap, or the logits themselves, using an explicit regularization term, will this have the desired effect of crushing the adversarial gradients?

\section{Easy ways to imitate adversarial training:  label smoothing \& logit squeezing}
There are two approaches to replicating the effect on the logits produced by adversarial training.  The first is to replicate the decrease in logit gap seen in Figure \ref{fig:mnist_eps_natNadv}.  This can be achieved by label smoothing.  A second approach to replicating adversarial training is to just directly crush all logit values and mimic the behavior in Figure \ref{fig:logit_dist_c10Nmnist}.  This approach is known as ``logit squeezing,'' and works by adding a regularization term to the training objective that explicitly penalizes large logits.

\paragraph{Label smoothing}
Label smoothing converts ``one-hot" label vectors into ``one-warm''  vectors that represents a low-confidence classification.  Because large logit gaps produce high-confidence classifications, label-smoothed training data forces the classifier to produce small logit gaps.  Label smoothing is a commonly used trick to prevent over-fitting on general classification problems, and it was first observed to boost adversarial robustness by \cite{warde201611}, where it was used as an inexpensive replacement for the network distillation defense \citep{papernot2016distillation}.
A one-hot label vector $\by_{hot}$ is smoothed using the formula
 $$\by_{warm} = \by_{hot} - \alpha \times (\by_{hot} - \frac{1}{N_{c}}),$$
  where $\alpha\in[0,1]$ is the smoothing parameter, and $N_c$ is the number of classes.  If we pick $\alpha=0$ we get a hard decision vector with no smoothing, while $\alpha=1$ creates an ambiguous decision by assigning equal certainty to all classes. 

\paragraph{Logit squeezing}
A second approach to replicating adversarial training is to just directly crush all logit values and mimic the behavior in Figure \ref{fig:logit_dist_c10Nmnist}.  This approach is known as ``logit squeezing,'' and works by adding a regularization term to the training objective that explicitly penalizes large logits.
 \cite{kannan2018adversarial} were the first to introduce logit-squeezing as an alternative to a ``logit pairing'' defense. 
Logit squeezing relies on the loss function
\begin{equation}
   \underset{\btheta}{\mathrm{minimize}} 
   \quad
 \sum_k L(\btheta, X_k, Y_k) + \beta ||\bz(X_k)||_F,
    \label{eq:logit_squeezing}
\end{equation}
where $\beta$ is the squeezing parameter (i.e., coefficient for the logit-squeezing term) and $||.||_F$ is the Frobenius norm of the logits for the mini-batch.

\paragraph{Can such simple regularizers really replicate adversarial training?} Our experimental results suggest that simple regularizers can {\em hurt} adversarial robustness, which agrees with the findings in~\cite{zantedeschi2017efficient}. 
However, these strategies become highly effective when combined with a simple trick from the adversarial training literature --- data augmentation with Gaussian noise.

\section{Gaussian noise saves the day}

Adding Gaussian noise to images during training (i.e, Gaussian noise augmentation) can be used  to improve the adversarial robustness of classifiers \citep{kannan2018adversarial, zantedeschi2017efficient}. However,
the effect of Gaussian noise is not well understood \citep{kannan2018adversarial}.  Here, we take a closer look at the behavior of Gaussian augmentation through systematic experimental investigations, and see that its effects are more complex than one might think.

\subsection{Gaussian noise synergistically interacts with regularizers}
Label smoothing and logit squeezing become shockingly effective at hardening networks when they are combined with Gaussian noise augmentation.  From the robustness plots in Figure \ref{fig:mnist_label_smooth_vs_rand}, we can see that training with Gaussian noise alone produces a noticeable change in robustness, which seems to be mostly attributable to a widening of the logit gap and slight decrease in the gradient gap ($\lVert \nabla_x z_y - \nabla_x z_{\bar{y}} \rVert_1$).   The small increase in robustness from random Gaussian augmentation was also reported by \cite{kannan2018adversarial}.   We also see that label smoothing alone causes a very slight drop in robustness; the shrink in the gradient gap is completely offset by a collapse in the logit gap.  

Surprisingly, Gaussian noise and label smoothing have a powerful synergistic effect.  When used together they cause a dramatic drop in the gradient gap, leading to a surge in robustness.  A similar effect happens in the case of logit squeezing, and results are shown in \cref{appendix:logit_squeeze} (Figure \ref{fig:mnist_logit_squeeze_vs_rand}).

\subsection{Gaussian noise helps regularization properties generalize ``off the manifold''}
Regularization methods have the potential to squash or align the adversarial gradients, but these properties are only imposed during training on images from the manifold that the ``true'' data lies on.  At test time, the classifier sees adversarial images that do not  ``look like'' training data because they lie off of, but adjacent to, the image manifold.  By training the classifier on images with random perturbations, we teach the classifier to enforce the desired properties for input images that lie off the manifold.

The generalization property of Gaussian augmentation seems to be independent from, and sometimes conflicting with, the synergistic properties discuss above.
In our experiments below, we find that smaller noise levels lead to a stronger synergistic effect, and yield larger $\epsilon_L$ and better robustness to FGSM attacks.  However, larger noise levels enable the regularization properties to generalize further off the manifold, resulting in better robustness to iterative attacks or attacks that escape the flattened region by adding an initial random perturbation.  See the results on MNIST in \cref{tab:mnist_logit_squeeze_and_label_smooth_abstractV} and the results on CIFAR-10 in \cref{tab:c10_white_Vabs} for various values of $\sigma$ (standard deviation of Gaussian noise). For more comprehensive experiments on the different parameters that contribute to the regularizers see Table \ref{tab:mnist_logit_squeeze_and_label_smooth_fullV} for MNIST and Tables \ref{tab:c10_white} \& \ref{tab:c10_black} in Appendices \ref{appendix:mnist_res} \& \ref{appendix:cifar_res}.

\epspack{mnist}{label_smooth}{Label smoothing by itself worsens robustness. With the addition of Gaussian augmentation, robustness improves. This enhanced robustness results from both gradients getting squashed and gradients becoming more aligned -- two effects that shrink the denominator in Equation \ref{eq:eps_smallest}. Plots use MNIST test samples with $\sigma=0.5$ and $\alpha=0.75$. The y-axis of the logit gap is chopped at 30k.}{fig:mnist_label_smooth_vs_rand}

\subsection{What is the difference between logit squeezing and label smoothing}
\begin{figure}[ht]
    \centering
    \begin{subfigure}[b]{0.49\textwidth}
        \includegraphics[width=\textwidth]{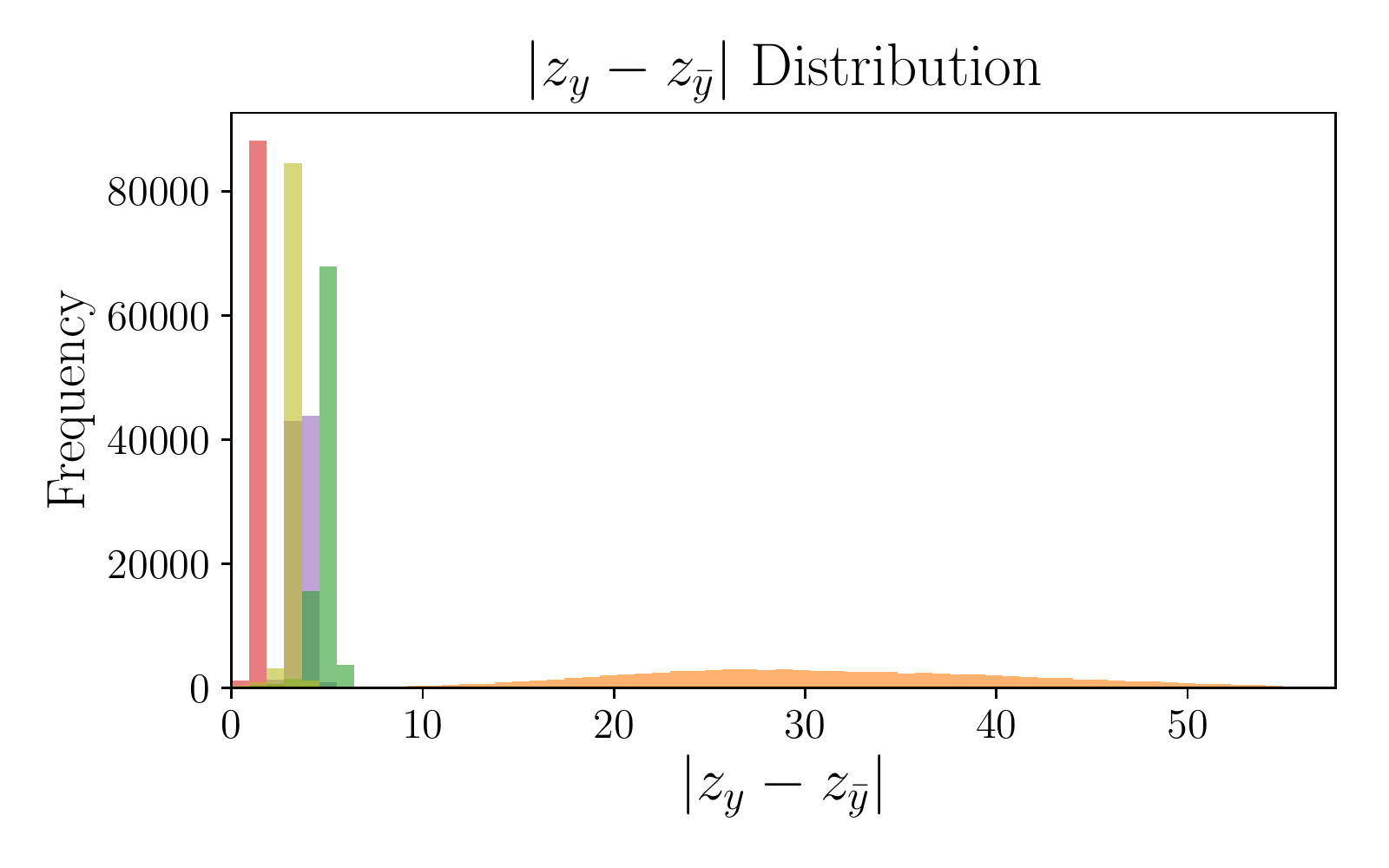}
    \end{subfigure}
    \begin{subfigure}[b]{0.49\textwidth}
        \includegraphics[width=\textwidth]{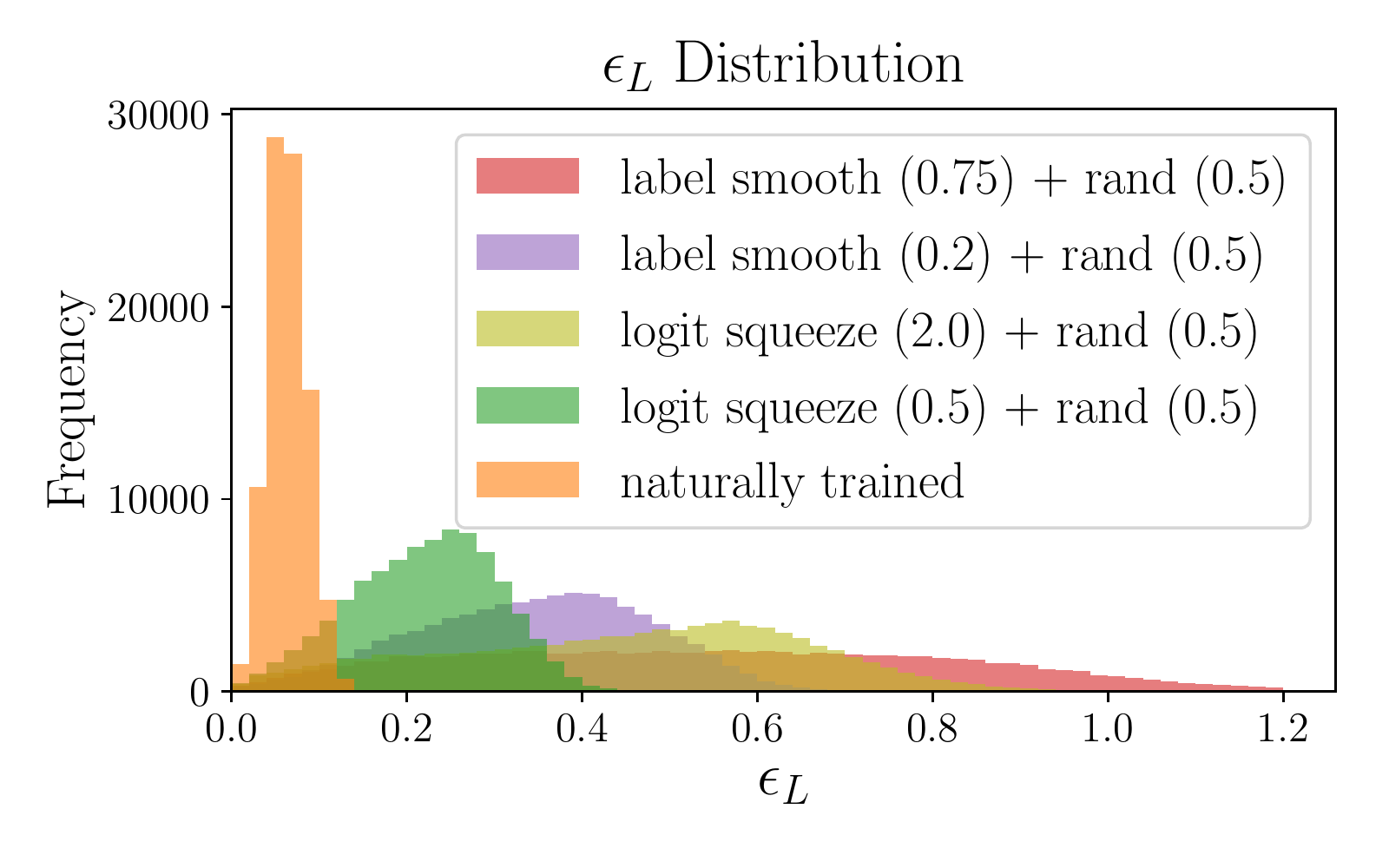}
    \end{subfigure}
        \centering
    \begin{subfigure}[b]{0.49\textwidth}
        \includegraphics[width=\textwidth]{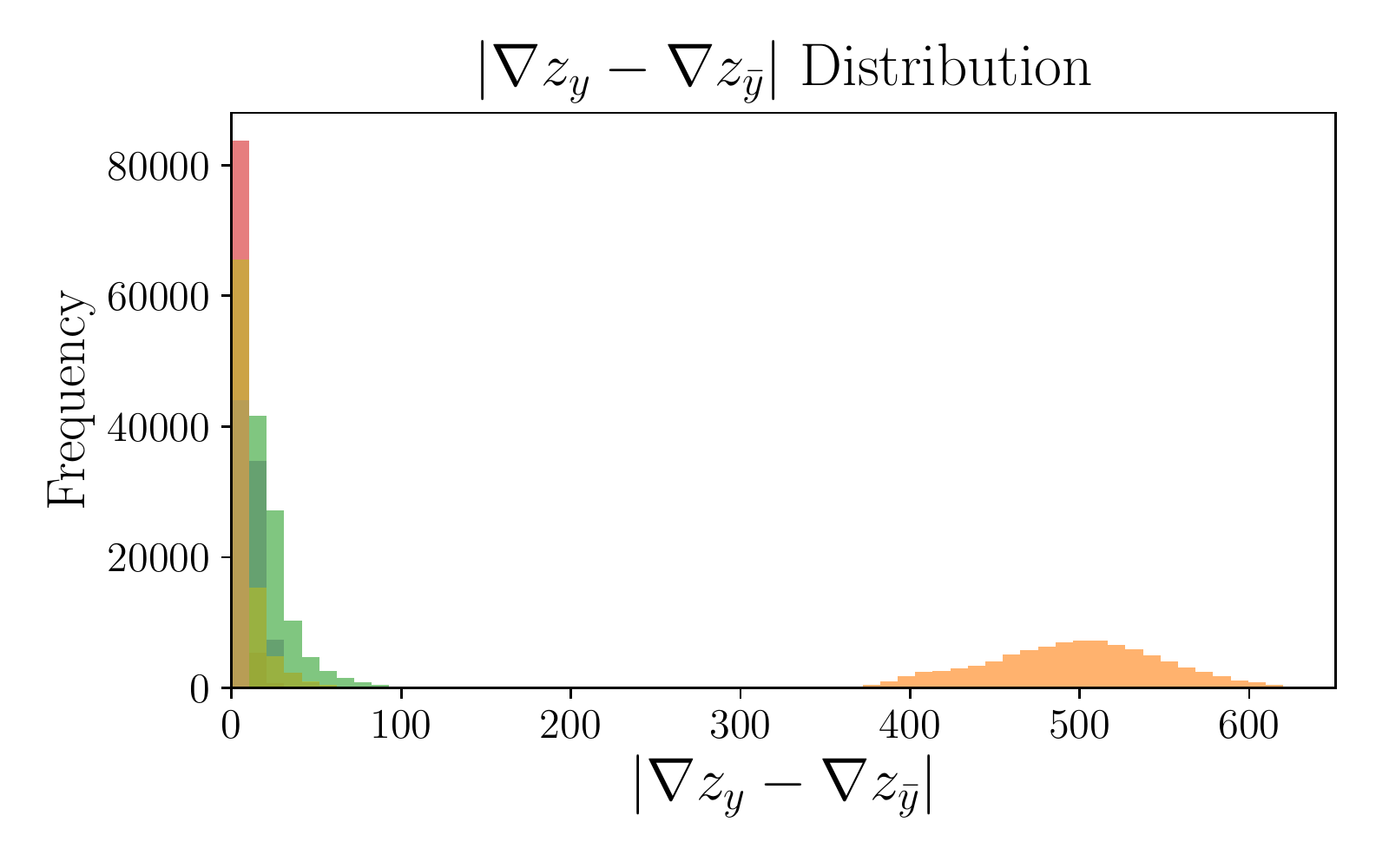}
    \end{subfigure}
    \begin{subfigure}[b]{0.49\textwidth}
        \includegraphics[width=\textwidth]{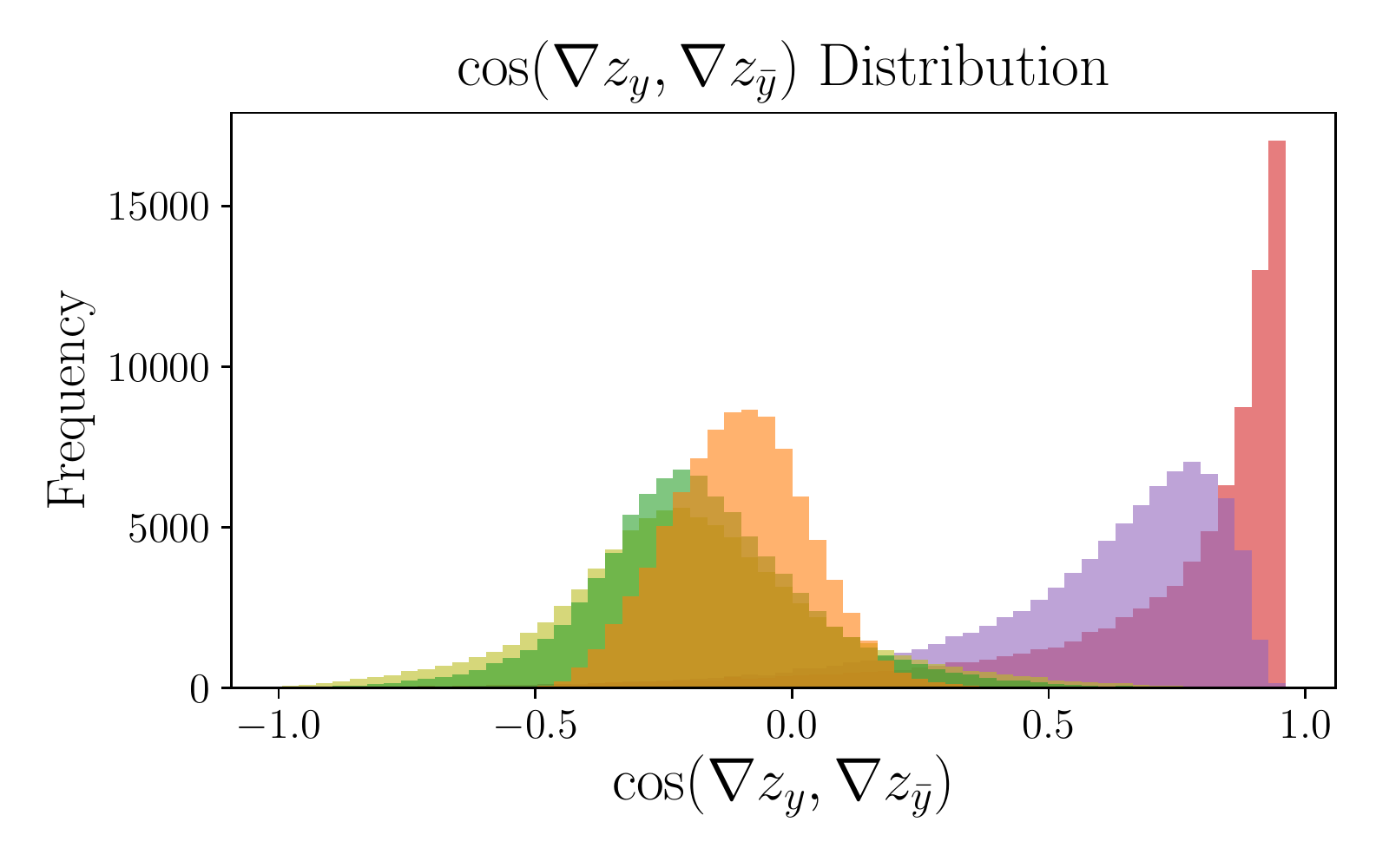}
    \end{subfigure}
    \caption{Label smoothing plus Gaussian augmentation works by decreasing the gradient gap. This is done by both shrinking the gradients and aligning them. The effects get larger with the increase in the smoothing parameter. Logit squeezing plus Gaussian noise works by increasing the logit gap and decreasing the gradient gap. Increasing the squeezing parameter, strengthens these effects.}
    \label{fig:different_logsq_labsm}
\end{figure}

Label smoothing (i.e. reducing the variance of the logits) is helpful because it causes the gradient gap to decrease.
The decreased gradient gap may be due to smaller element-wise gradient amplitudes, the alignments of the adversarial gradients, or both. To investigate the causes,  we plot the $\ell_1$ norm of the gradients of the logits with respect to the input image \footnote{This plot is moved to the appendix due to space limitations. See \cref{appendix:logit_squeeze} (\cref{fig:1norm_grads_dist}).} and the cosine of the angle between the gradients (\cref{fig:different_logsq_labsm}). We see that in label smoothing (with Gaussian augmentation), both the gradient magnitude decreases and the gradients get more aligned. Larger smoothing parameters $\alpha$ cause the gradient to be both smaller and more aligned.

When logit squeezing is used with Gaussian augmentation, the magnitudes of the gradients decrease. The distribution of the cosines between gradients widens, but does not increase like it did for label smoothing.  These effects are very similar to the behavior of adversarial training in \cref{fig:mnist_eps_natNadv}. 
Interestingly, in the case of logit squeezing with Gaussian noise, unlike label smoothing, the numerator of Equation \ref{eq:eps_smallest} increases as well. 
This increase in the logit gap disappears once we take away Gaussian augmentation (See \cref{appendix:logit_squeeze} \cref{fig:mnist_logit_squeeze_vs_rand}). Simultaneously increasing the numerator and decreasing the denominator of Equation \ref{eq:eps_smallest} potentially gives a slight advantage to logit squeezing.

\subsection{MNIST results for various defense parameters}
There are multiple factors that can affect the robustness of the MNIST classifier \footnote{Similar to other experiments, we use the same architecture and hyper-parameters of \cite{madry2017towards}}. While regularizers do not yield more robustness than adversarial training for MNIST, the results are promising given that these relatively high values of robustness come at a cheap cost in comparison to adversarial training. In Table \ref{tab:mnist_logit_squeeze_and_label_smooth_abstractV} we notice that as we increase the number of training iterations $k$, we get more robust models for both logit squeezing and label smoothing \footnote{For comprehensive results with more sensitivity analysis on the parameters, see \cref{tab:mnist_logit_squeeze_and_label_smooth_fullV} in \cref{appendix:mnist_res}.}. We get more robust models when we use larger smoothing ($\alpha$) and squeezing ($\beta$) parameters, and when Gaussian augmentation is used with standard deviation $\sigma$ that is greater than the desired $\epsilon$ (the maximum perturbation size).

\begin{table}[ht]
    \centering
    \begin{tabular}{c|c|c|c|c|c|c|c|c|c}
    \multicolumn{4}{c}{defense params} & & & \multicolumn{2}{c}{White-box} & \multicolumn{2}{c}{black-box}\\
    & & & & &  &\multicolumn{2}{c}{40-step PGD} &  & \\
    $\alpha$ & $\beta$ & $\sigma$ & $k$ & Train &  Test & X-ent & CW & FGSM & PGD\\
    \hline
    0 & 0.5 & 0.5 & 400k & 99.96\% & 99.21\% & 72.44\% & 72.39\%& 87.58\% & 89.48\%\\
    0 & 2.0 & 0.0 & 400k & 63.11\% & 64.14\% & 0.00\% & 0.00\% & 17.69\% & 24.52\%\\
    0 & 2.0 & 0.3 & 400k & 99.99\% & 97.17\% & 76.08\% & 76.28\% & 82.11\% & 85.80\%\\
    0 & 2.0 & 0.5 & 400k & 99.94\% & 99.21\% & 79.13\% & 78.36\% & 87.75\% & 89.47\%\\
    0 & 5.0 & 0.5 & 400k & 99.84\% & 99.18\% & 78.21\% & 77.29\% & 87.73\% & 89.25\%\\
    \hline
    0.2 & 0 & 0 & 400k & 100.00\% & 99.36\% & 0.00\% & 0.00\% & 23.88\% & 29.33\%\\
    0.2 & 0 & 0.1 & 400k & 100.00\% & 99.39\% & 15.59\% & 17.91\% & 59.14\% & 63.73\%\\
    0.2 & 0 & 0.3 & 400k & 100.00\% & 99.40\% & 70.97\% & 73.71\% & 81.60\% & 85.38\%\\
    0.2 & 0 & 0.7 & 400k & 99.76\% & 99.04\% & 66.34\% & 70.96\% & 87.27\% & 88.60\%\\
    0.95 & 0 & 0.3 & 100k & 99.78\% & 99.38\% & 60.70\% & 61.93\% & 94.29\% & 77.21\%\\
    0.95 & 0 & 0.3 & 400k & 99.99\% & 99.44\% & 74.02\% & 75.46\% & 93.97\% & 85.00\%\\
    0.95 & 0 & 0.3 & 2M & 100.00\% & 99.25\% & 83.60\% & 85.39\% & 85.39\% & 87.22\%\\
    \hline
    
    \multicolumn{4}{c}{\cite{madry2017towards} adv tr.} & 100\% & 98.8\% & 93.20\% & 93.9 \% & 96.08\% & 96.81\%\\ 
    \end{tabular}
    \caption{Accuracy of different MNIST classifiers against PGD and FGSM attacks on X-ent and CW losses under the white-box and black-box threat models. Attacks have maximum $\ell_\infty$ perturbation $\epsilon=0.3$. The iterative white-box attacks have an initial random step. The naturally trained model is used for generating black-box attacks. We use CW loss for the black-box attack.}
    \label{tab:mnist_logit_squeeze_and_label_smooth_abstractV}
\end{table}

\section{Aggressive Label Smoothing and Logit squeezing can outperform adversarial training on Cifar-10} \label{sec:cifar10res}
We trained Wide-Resnet CIFAR-10 classifiers (depth=28 and k=10 ) using aggressive values for the smoothing and squeezing parameters on the CIFAR10 data set. Similar to \cite{madry2017towards}, we use the standard data-augmentation techniques and weight-decay. We compare our results to those of \cite{madry2017towards}. Note that the adversarially trained model from \cite{madry2017towards} has been trained for 80,000 iterations on adversaries which are generated using a 7-step PGD. Keeping in mind that each step requires a forward and backward pass, the running time of training for 80,000 iterations on 7-step PDG examples is equivalent to 640,000 iterations of training with label smoothing or logit squeezing. A short version of our results on white-box attacks are summarized in Table \ref{tab:c10_white_Vabs}.  The results of some of our black-box experiments are summarized in Table \ref{tab:c10_black_Vabs}\footnote{For more complete results see \cref{tab:c10_white} and \cref{tab:c10_black} in \cref{appendix:cifar_res}.}. While logit squeezing seems to outperform label smoothing in the white-box setting, label smoothing is slightly better under the black-box threat.

We see that aggressive logit squeezing with squeezing parameter $\beta=10$ and $\sigma=20$ results in a model that is more robust than the adversarially trained model from \cite{madry2017towards} when attacked with PGD-20. Interestingly, it also achieves higher test accuracy on clean examples. 
\begin{table}[!h]
    \centering
    \begin{tabular}{c|c|c|c|c|c|c|c|c|c}
    \multicolumn{4}{c}{defense params} & \multicolumn{6}{c}{White-box}\\
    & & & & & &\multicolumn{2}{c}{20-step PGD + Rand} & \multicolumn{2}{c}{FGSM} \\
    $\alpha$ & $\beta$ & $\sigma$ & k & Train &  Test & xent & CW & xent & CW \\
    \hline
    0.95 & 0 & 20 & 160k & 99.90\% & \bf{92.88\%} & 43.00\% & 41.29\% & 75.25\% & 74.51\%  \\
    0.95 & 0 & 30 & 160k & 99.64\% & 90.70\% & \bf{53.93\%} & 40.68\% & 64.77\% & 70.38\% \\
    0.95 & 0 & 30 & 240k & 99.70\% & 90.55\% & 47.27\% & 37.25\% & 64.36\% & 69.44\%  \\
    0.8 & 0 & 30 & 320k & 99.72\% & 90.23\% & 43.51\% & 42.96\% & 74.68\% & 72.66\%  \\
    \hline
    0 & 10 & 20 & 160k & 99.92\% & 92.68\% & 52.55\% & 48.78\% & \bf{76.37\%} & \bf{75.79\%} \\ 
    0 & 10 & 30 & 80k & 99.45\% & 89.89\% & 48.46\% & 45.51\% & 68.19\% & 67.25\% \\ 
    0 & 10 & 30 & 160k & 99.82\% & 90.49\% & 52.30\% & \bf{49.73\%} & 72.08\% & 71.08\% \\ 
    \hline
    \multicolumn{4}{c}{\cite{madry2017towards}} & \bf{100.00\%} & 87.25\% & 45.84\% & 46.90\% & 56.22\% & 55.57\% \\

    \end{tabular}
    \caption{White-box attacks on CIFAR-10 models. We use $\ell_{\infty}$ attacks with $\epsilon=8$. For the 20-step PGD, similarly to \cite{madry2017towards}, we use an initial random perturbation. We do not use a random perturbation for the FGSM attack since it decreased the attack's effectiveness.
    }
    \label{tab:c10_white_Vabs}
\end{table}

\begin{table}[!h]
    \centering
    \begin{tabular}{c|c|c|c|c|c|c|c|c|c}
    \multicolumn{4}{c}{defense params} & \multicolumn{6}{c}{Black-box}\\
    & & & & \multicolumn{2}{c}{7-step PGD} &\multicolumn{2}{c}{7-step PGD + Rand} & \multicolumn{2}{c}{FGSM} \\
    $\alpha$ & $\beta$ & $\sigma$ & k & xent &  CW & xent & CW & xent & CW \\
    \hline
    0.95 & 0 & 20 & 160k & \bf{71.58\%} & \bf{71.96\%} & \bf{72.44\%} & \bf{73.16\%} & \bf{74.01\%} & \bf{74.68\%} \\
    0.95 & 0 & 30 & 160k & 68.33\% & 68.88\% & 69.09\% & 69.63\% & 70.85\% & 71.71\% \\
    0.95 & 0 & 30 & 240k & 67.88\% & 68.59\% & 68.84\% & 69.63\% & 70.53\% & 71.32\%  \\
    0.8 & 0 & 30 & 320k & 67.55\% & 68.32\% & 68.49\% & 69.28\% & 70.03\% & 70.89\%  \\
    \hline
    0 & 10 & 20 & 80k & 70.27\% & 70.78\% & 71.29\% & 71.86\% & 72.47\% & 73.30\% \\ 
    0 & 10 & 20 & 160k & 70.75\% & 71.49\% & 71.63\% & 72.22\% & 73.48\% & 73.82\%\\ 
    0 & 10 & 30 & 80k & 66.53\% & 67.46\% & 67.52\% & 68.40\% & 69.30\% & 69.99\% \\ 
    0 & 10 & 30 & 160k & 67.05\% & 67.79\% & 68.30\% & 68.89\% & 69.94\% & 70.61\%\\ 
    \hline
    \multicolumn{4}{c}{\cite{madry2017towards}} & 63.39\%* & 64.38\%* & 63.39\%* & 64.38\%* & 67.00\% & 67.25\% \\

    \end{tabular}
    \caption{Black-box attacks on CIFAR-10 models. Attacks are $\ell_{\infty}$ with $\epsilon=8$. Similar to \cite{madry2017towards}, We build 7-step PGD attacks and FGSM attacks for a public adversarially trained model. *Values taken from the original paper by \cite{madry2017towards}.}
    \label{tab:c10_black_Vabs}
\end{table}

\subsection{Attacking with stronger adversaries}
Some defenses work by obfuscating gradients and masking the gradients. \cite{DBLP:journals/corr/abs-1802-00420} suggest these models can be identified by performing ``sanity checks'' such as attacking them with unbounded strong adversaries (\textit{i.e.} unbounded $\epsilon$ with many iterations). By attacking our robust models using these unbounded attacks, we verify that the unbounded adversary can degrade the accuracy to 0.00\% which implies that the adversarial example generation optimization attack (PGD) is working properly.
Also, it is known that models which do not completely break the PGD attack (such as us) can possibly lead to a false sense of security by creating a loss landscape that prevents an $\epsilon$-bounded but weak adversary from finding strong adversarial examples. This can be done by convoluting the loss landscape such that many local-optimal points exist. This false sense of security can be identified by increasing the strength of the adversary using methods such as increasing the number of steps and random restarts of the PGD attack. We run the stronger PGD attacks on a sample model from \cref{tab:c10_white_Vabs} with hyper-parameters $k=160k, \beta=10$, and $\sigma=30$. We notice that performing 9 random restarts for the PGD attack on the cross-entropy loss only drops the accuracy slightly to 49.86\%. Increasing the number of PGD steps to 1000 decreases the accuracy slightly more to 40.94\%
\footnote{See \cref{tab:num_restarts} and \cref{tab:pgd_iteration_accuracy} in the appendix for more complete results.}. 
While for such strong iterative white-box attacks our sample model is less robust than the adversarially trained model, there are other areas where this hardened model is superior to the adversarially trained model: Very high robustness in the black box setting (roughly 3\% higher than that for adversarial training according to \cref{tab:c10_black_Vabs}) and against white-box non-iterative (or less iterative) attacks (roughly 15\%); 
high test accuracy on clean data (roughly 3\%); and,
very fast training time compared to adversarial training.

\subsection{The loss landscape of a logit-squeezed model for CIFAR-10}
\begin{figure}
    \centering
    \includegraphics[width=\textwidth]{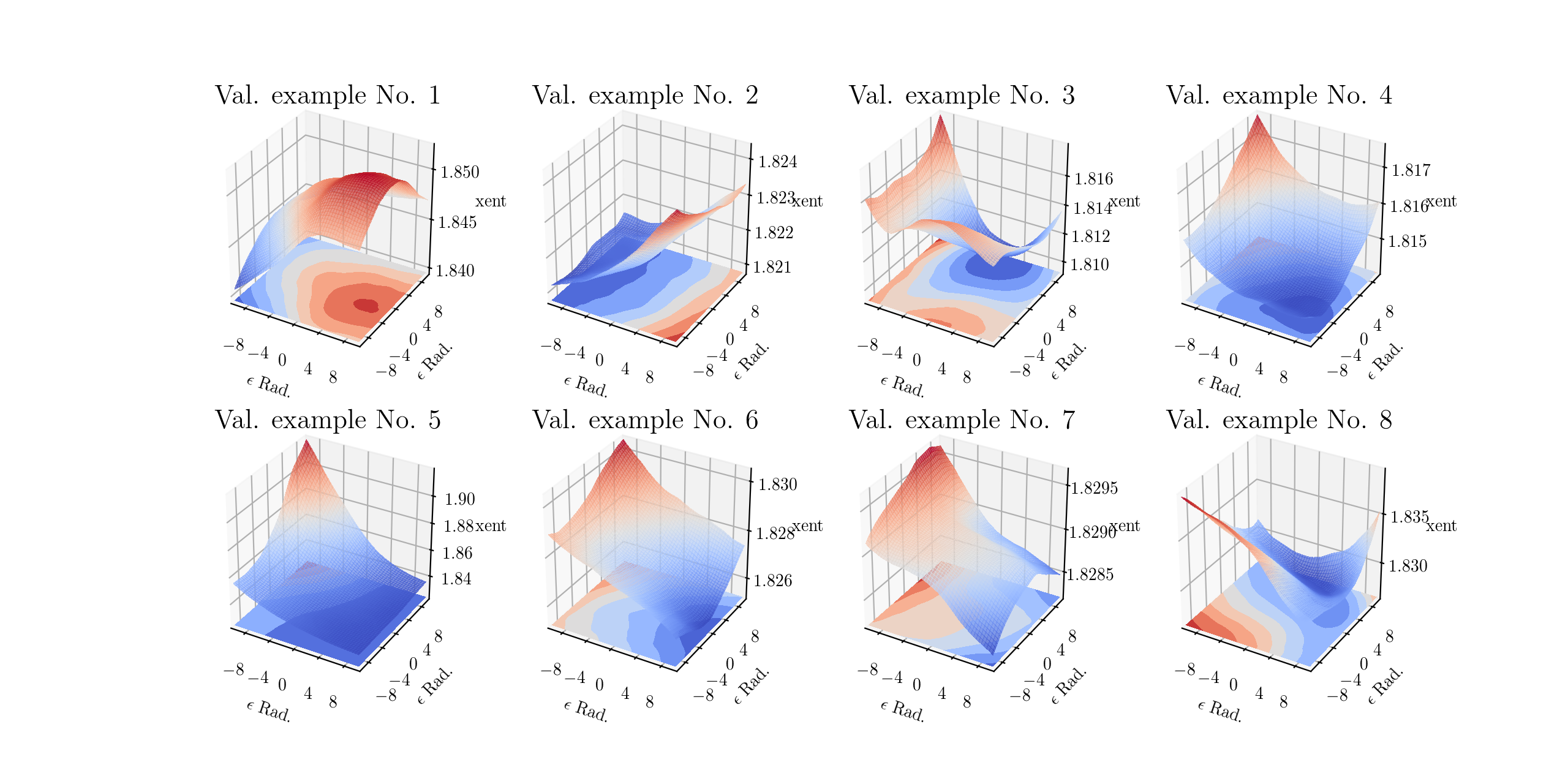}
    \caption{The cross-entropy landscape of the first eight images of the validation set for the model trained for 160k iterations with hyper-parameters $\beta=10$ and $\sigma=30$.
    To plot the loss landscape we take two random directions $r_1$ and $r_2$ (\textit{i.e.} $r_1$, $r_2 $
    $\sim$ 
    Rademacher(0.5)
    ). We plot the cross-entropy (\textit{i.e. xent}) loss at different points $x = x_i + \epsilon_1 \cdot r_1 + \epsilon_2 \cdot r_2$. Where $x_i$ is the cletan image and $  -10\leq \epsilon_1, \epsilon_2 \leq 10$.
    }
    \label{fig:rad_drections}
\end{figure}

To further verify that our robust model is not falsely creating a sense of robustness by ``breaking'' the optimization problem that generates adversarial examples by either masking the gradients or making the loss landscape convoluted, we visualize the loss landscape of our sample model from \cref{tab:c10_white_Vabs}. We plot the classification (\textit{e.g.,} cross-entropy) loss for points surrounding the first eight validation images that belong to the subspace spanned by two random directions\footnote{See \cref{fig:rad_adv_directions} in \cref{appendix:loss_land} for the xent loss landscape along the adversarial and random directions.} in \cref{fig:rad_drections}. It seems that the loss landscape has not become convoluted. This observation is backed up by the fact that increasing the number of PGD attack steps does not substantially affect accuracy.

\section{Hardening CIFAR-100 classifiers with aggressive logit squeezing}
To check the performance of our proposed regularizers on more complicated datasets with more number of classes, we perform aggressive logit squeezing on the CIFAR-100 dataset which contains 100 categories. We use the same architecture and settings used for training the CIFAR-10 classifiers. The white-box performance of two hardened models with logit squeezing and a PGD adversarially trained model for the same architecture are summarized in \cref{tab:c100_white_Vabs}.

\begin{table}[!h]
    \centering
    \begin{tabular}{c|c|c|c|c|c|c|c|c|c}
    \multicolumn{4}{c}{defense params} & \multicolumn{6}{c}{White-box}\\
    $\alpha$ & $\beta$ & $\sigma$ & k & Train &  Test & PGD-7 & PGD-20 & PGD-100 & PGD-200 \\
    \hline
    \hline
    0 & 5 & 20 & 80k & 98.99\% & 69.80\% & 30.68\% & 22.91\% & 18.69\% & 18.09\% \\ 
    0 & 5 & 20 & 160k & 99.42\% & \textbf{70.21\%} & \textbf{37.08\%} & \textbf{30.91\%} & \textbf{26.47\%} & \textbf{26.00\%} \\ 
    \hline
    \multicolumn{4}{c}{PGD-2 adv trained}  & \textbf{99.96\%} & 67.94\% & 20.08\% & 17.08\% & 16.49\% & 16.50\% \\

    \end{tabular}
    \caption{White-box iterative attacks on CIFAR-100 models. We use $\ell_{\infty}$ attacks with $\epsilon=8$. For brevity, we only report the results for attacking the cross-entropy loss. We attack the models with adversaries having different strengths by varying the number of PGD steps.
    }
    \label{tab:c100_white_Vabs}
\end{table}

Similar to CIFAR-10, aggressive logit squeezing can result in models as robust as those that are adversarially trained at a fraction of the cost. The logit-squeezed model that is trained for only 80k iterations achieves high classification accuracy for natural/clean examples. It is also more robust against white-box PGD attacks compared to the adversarially trained model that requires much more training time. The logit-squeezed model with $k=160$k improves the robustness and clean accuracy even further and still trains faster than the adversarially trained model (28.8 hours vs 34.3 hours).

\section{conclusion}
We studied the robustness of adversarial training, label smoothing, and logit squeezing through a linear approximation $\epsilon_L$ that relates the magnitude of adversarial perturbations to the logit gap and the difference between the adversarial directions for different labels.
Using this simple model, we observe how adversarial training achieves robustness and try to imitate this robustness using label smoothing and logit squeezing. The resulting methods perform well on MNIST, and can get results on CIFAR-10 and CIFAR-100 that can excel over adversarial training in both robustness and accuracy on clean examples.  By demonstrating the effectiveness of these simple regularization methods, we hope this work can help make robust training easier and more accessible to practitioners.

\bibliography{iclr2019_conference}

\begin{thebibliography}{21}
\providecommand{\natexlab}[1]{#1}
\providecommand{\url}[1]{\texttt{#1}}
\expandafter\ifx\csname urlstyle\endcsname\relax
  \providecommand{\doi}[1]{doi: #1}\else
  \providecommand{\doi}{doi: \begingroup \urlstyle{rm}\Url}\fi

\bibitem[Athalye et~al.(2018)Athalye, Carlini, and
  Wagner]{DBLP:journals/corr/abs-1802-00420}
Anish Athalye, Nicholas Carlini, and David~A. Wagner.
\newblock Obfuscated gradients give a false sense of security: Circumventing
  defenses to adversarial examples.
\newblock \emph{CoRR}, abs/1802.00420, 2018.
\newblock URL \url{http://arxiv.org/abs/1802.00420}.

\bibitem[Biggio et~al.(2013)Biggio, Corona, Maiorca, Nelson, {\v{S}}rndi{\'c},
  Laskov, Giacinto, and Roli]{biggio2013evasion}
Battista Biggio, Igino Corona, Davide Maiorca, Blaine Nelson, Nedim
  {\v{S}}rndi{\'c}, Pavel Laskov, Giorgio Giacinto, and Fabio Roli.
\newblock Evasion attacks against machine learning at test time.
\newblock In \emph{Joint European conference on machine learning and knowledge
  discovery in databases}, pp.\  387--402. Springer, 2013.

\bibitem[Carlini \& Wagner(2017)Carlini and Wagner]{carlini2017towards}
Nicholas Carlini and David Wagner.
\newblock Towards evaluating the robustness of neural networks.
\newblock In \emph{2017 IEEE Symposium on Security and Privacy (SP)}, pp.\
  39--57. IEEE, 2017.

\bibitem[Evtimov et~al.(2017)Evtimov, Eykholt, Fernandes, Kohno, Li, Prakash,
  Rahmati, and Song]{evtimov2017robust}
Ivan Evtimov, Kevin Eykholt, Earlence Fernandes, Tadayoshi Kohno, Bo~Li, Atul
  Prakash, Amir Rahmati, and Dawn Song.
\newblock Robust physical-world attacks on deep learning models.
\newblock \emph{arXiv preprint arXiv:1707.08945}, 1, 2017.

\bibitem[Girshick(2015)]{girshick2015fast}
Ross Girshick.
\newblock Fast r-cnn.
\newblock In \emph{Proceedings of the IEEE international conference on computer
  vision}, pp.\  1440--1448, 2015.

\bibitem[Goodfellow et~al.(2014)Goodfellow, Shlens, and
  Szegedy]{goodfellow2014explaining}
Ian~J Goodfellow, Jonathon Shlens, and Christian Szegedy.
\newblock Explaining and harnessing adversarial examples.
\newblock \emph{arXiv preprint arXiv:1412.6572}, 2014.

\bibitem[Kannan et~al.(2018)Kannan, Kurakin, and
  Goodfellow]{kannan2018adversarial}
Harini Kannan, Alexey Kurakin, and Ian Goodfellow.
\newblock Adversarial logit pairing.
\newblock \emph{arXiv preprint arXiv:1803.06373}, 2018.

\bibitem[Krizhevsky et~al.(2012)Krizhevsky, Sutskever, and
  Hinton]{krizhevsky2012imagenet}
Alex Krizhevsky, Ilya Sutskever, and Geoffrey~E Hinton.
\newblock Imagenet classification with deep convolutional neural networks.
\newblock In \emph{Advances in neural information processing systems}, pp.\
  1097--1105, 2012.

\bibitem[Kurakin et~al.(2016)Kurakin, Goodfellow, and
  Bengio]{kurakin2016adversarialBIM}
Alexey Kurakin, Ian Goodfellow, and Samy Bengio.
\newblock Adversarial examples in the physical world.
\newblock \emph{arXiv preprint arXiv:1607.02533}, 2016.

\bibitem[Madry et~al.(2017)Madry, Makelov, Schmidt, Tsipras, and
  Vladu]{madry2017towards}
Aleksander Madry, Aleksandar Makelov, Ludwig Schmidt, Dimitris Tsipras, and
  Adrian Vladu.
\newblock Towards deep learning models resistant to adversarial attacks.
\newblock \emph{arXiv preprint arXiv:1706.06083}, 2017.

\bibitem[Meng \& Chen(2017)Meng and Chen]{meng2017magnet}
Dongyu Meng and Hao Chen.
\newblock Magnet: a two-pronged defense against adversarial examples.
\newblock In \emph{Proceedings of the 2017 ACM SIGSAC Conference on Computer
  and Communications Security}, pp.\  135--147. ACM, 2017.

\bibitem[Najibi et~al.(2017)Najibi, Samangouei, Chellappa, and
  Davis]{najibi2017ssh}
Mahyar Najibi, Pouya Samangouei, Rama Chellappa, and Larry~S Davis.
\newblock Ssh: Single stage headless face detector.
\newblock In \emph{ICCV}, pp.\  4885--4894, 2017.

\bibitem[Papernot et~al.(2016)Papernot, McDaniel, Wu, Jha, and
  Swami]{papernot2016distillation}
Nicolas Papernot, Patrick McDaniel, Xi~Wu, Somesh Jha, and Ananthram Swami.
\newblock Distillation as a defense to adversarial perturbations against deep
  neural networks.
\newblock In \emph{2016 IEEE Symposium on Security and Privacy (SP)}, pp.\
  582--597. IEEE, 2016.

\bibitem[Samangouei et~al.(2018)Samangouei, Kabkab, and
  Chellappa]{samangouei2018defense}
Pouya Samangouei, Maya Kabkab, and Rama Chellappa.
\newblock Defense-gan: Protecting classifiers against adversarial attacks using
  generative models.
\newblock \emph{arXiv preprint arXiv:1805.06605}, 2018.

\bibitem[Sharif et~al.(2016)Sharif, Bhagavatula, Bauer, and
  Reiter]{sharif2016accessorize}
Mahmood Sharif, Sruti Bhagavatula, Lujo Bauer, and Michael~K Reiter.
\newblock Accessorize to a crime: Real and stealthy attacks on state-of-the-art
  face recognition.
\newblock In \emph{Proceedings of the 2016 ACM SIGSAC Conference on Computer
  and Communications Security}, pp.\  1528--1540. ACM, 2016.

\bibitem[Shen et~al.(2017)Shen, Jin, Gao, and Zhang]{shen2017ape}
Shiwei Shen, Guoqing Jin, Ke~Gao, and Yongdong Zhang.
\newblock Ape-gan: Adversarial perturbation elimination with gan.
\newblock \emph{ICLR Submission, available on OpenReview}, 4, 2017.

\bibitem[Szegedy et~al.(2013)Szegedy, Zaremba, Sutskever, Bruna, Erhan,
  Goodfellow, and Fergus]{szegedy2013intriguing}
Christian Szegedy, Wojciech Zaremba, Ilya Sutskever, Joan Bruna, Dumitru Erhan,
  Ian Goodfellow, and Rob Fergus.
\newblock Intriguing properties of neural networks.
\newblock \emph{arXiv preprint arXiv:1312.6199}, 2013.

\bibitem[Tram{\`e}r et~al.(2017)Tram{\`e}r, Kurakin, Papernot, Goodfellow,
  Boneh, and McDaniel]{tramer2017ensemble}
Florian Tram{\`e}r, Alexey Kurakin, Nicolas Papernot, Ian Goodfellow, Dan
  Boneh, and Patrick McDaniel.
\newblock Ensemble adversarial training: Attacks and defenses.
\newblock \emph{arXiv preprint arXiv:1705.07204}, 2017.

\bibitem[Warde-Farley \& Goodfellow(2016)Warde-Farley and
  Goodfellow]{warde201611}
David Warde-Farley and Ian Goodfellow.
\newblock 11 adversarial perturbations of deep neural networks.
\newblock \emph{Perturbations, Optimization, and Statistics}, pp.\  311, 2016.

\bibitem[Xu et~al.(2017)Xu, Evans, and Qi]{xu2017feature}
Weilin Xu, David Evans, and Yanjun Qi.
\newblock Feature squeezing: Detecting adversarial examples in deep neural
  networks.
\newblock \emph{arXiv preprint arXiv:1704.01155}, 2017.

\bibitem[Zantedeschi et~al.(2017)Zantedeschi, Nicolae, and
  Rawat]{zantedeschi2017efficient}
Valentina Zantedeschi, Maria-Irina Nicolae, and Ambrish Rawat.
\newblock Efficient defenses against adversarial attacks.
\newblock In \emph{Proceedings of the 10th ACM Workshop on Artificial
  Intelligence and Security}, pp.\  39--49. ACM, 2017.

\end{thebibliography}
\bibliographystyle{iclr2019_conference}
\newpage
\appendix
\begin{center}{\Large Appendix: \mytitle}\end{center}

\section{Adversarial training effects on CIFAR-10}\label{appendix:cifar10_natNadv}
\begin{figure}[!htbp]
    \centering
    \begin{subfigure}[b]{0.49\textwidth}
        \includegraphics[width=\textwidth]{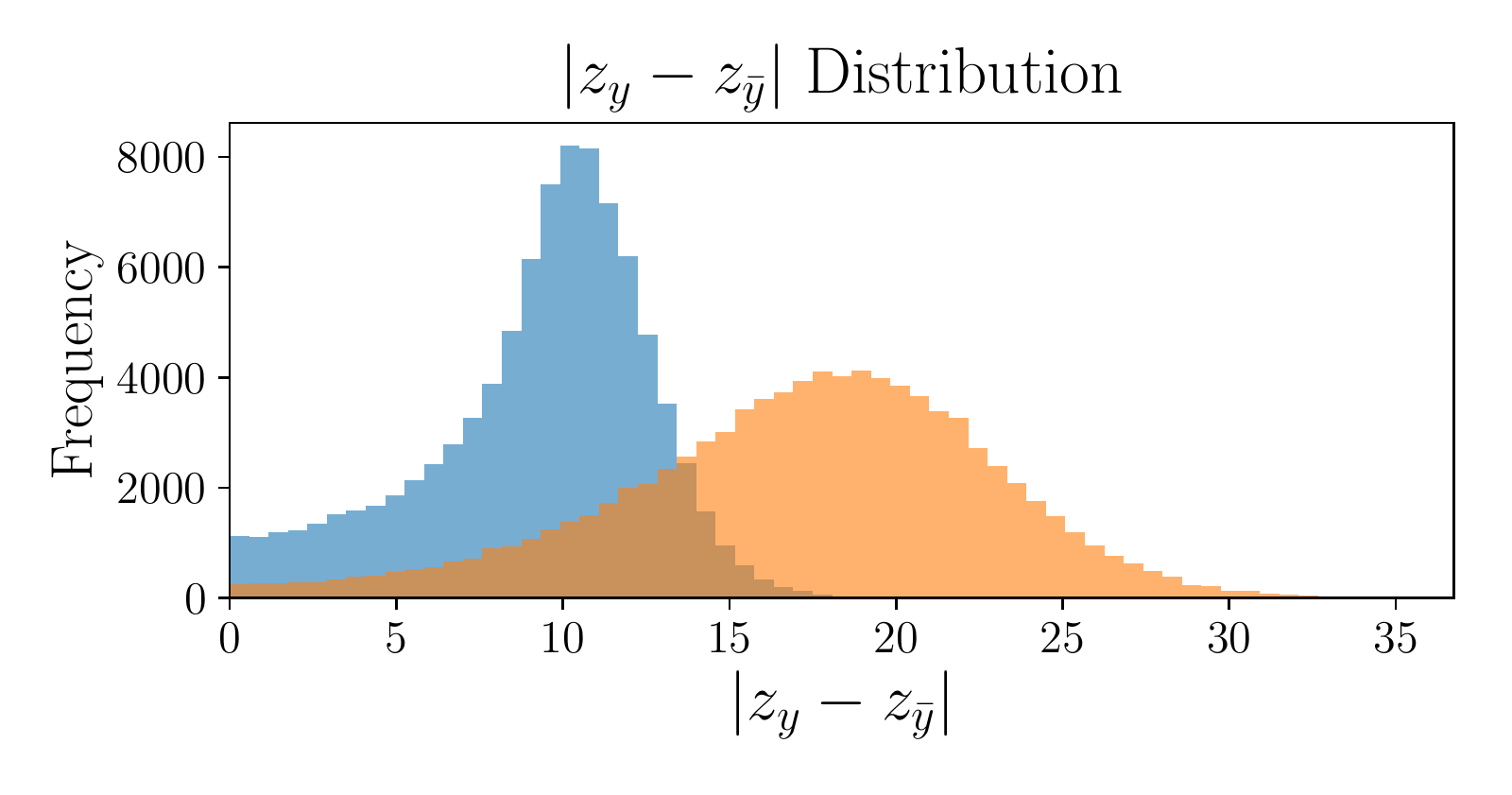}
    \end{subfigure}
    \begin{subfigure}[b]{0.49\textwidth}
        \includegraphics[width=\textwidth]{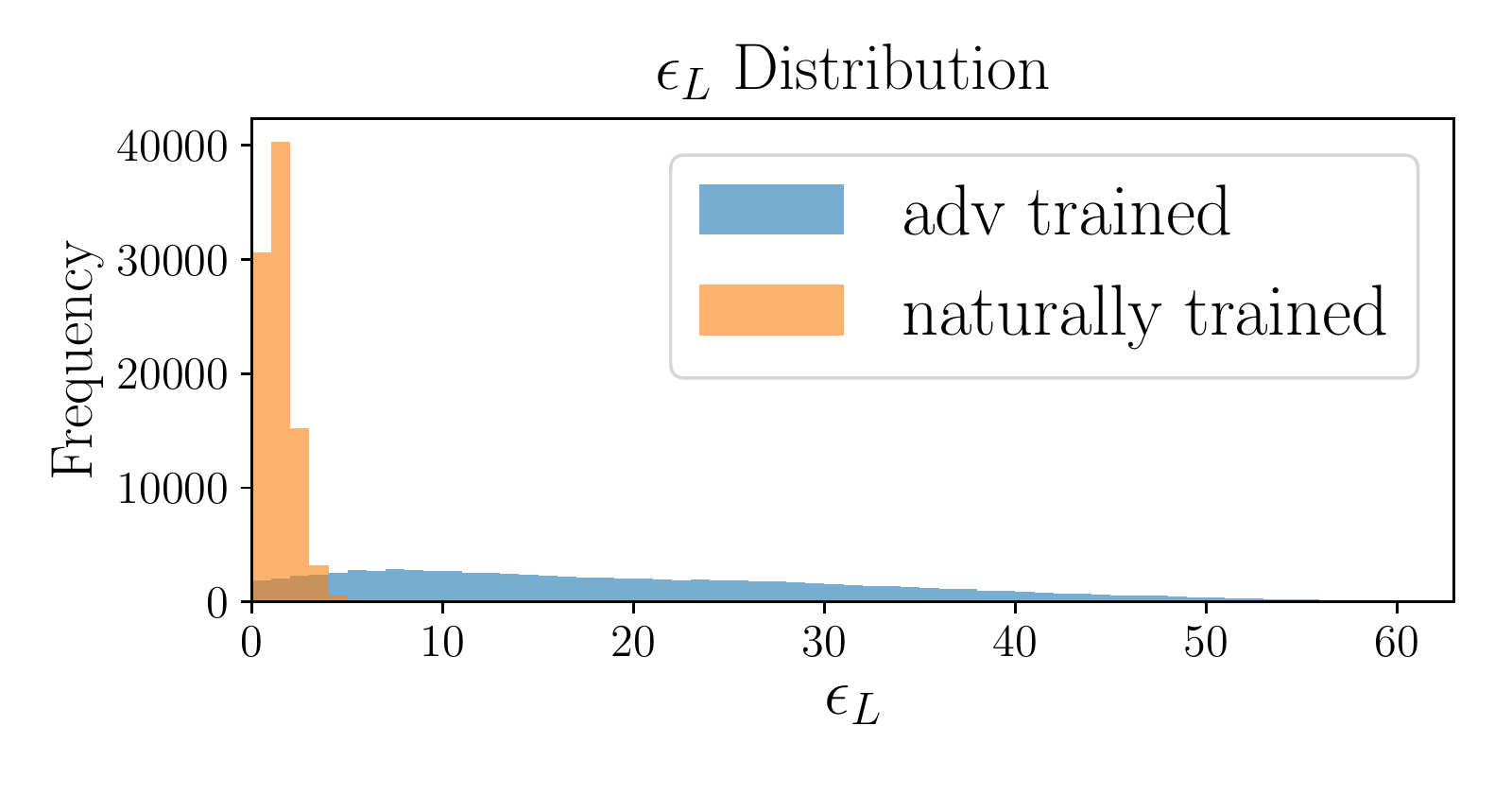}
    \end{subfigure}
        \centering
    \begin{subfigure}[b]{0.49\textwidth}
        \includegraphics[width=\textwidth]{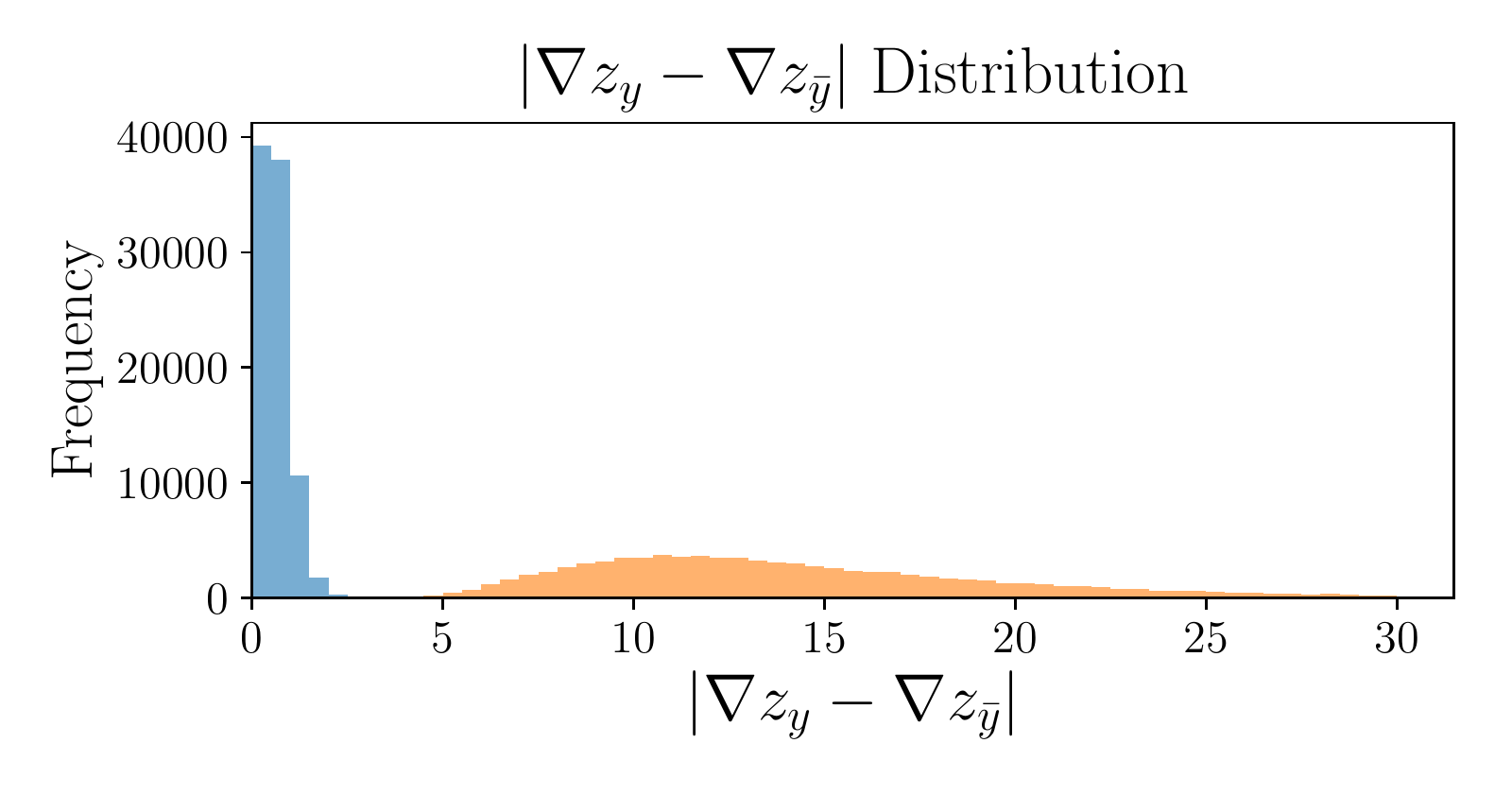}
    \end{subfigure}
    \begin{subfigure}[b]{0.49\textwidth}
        \includegraphics[width=\textwidth]{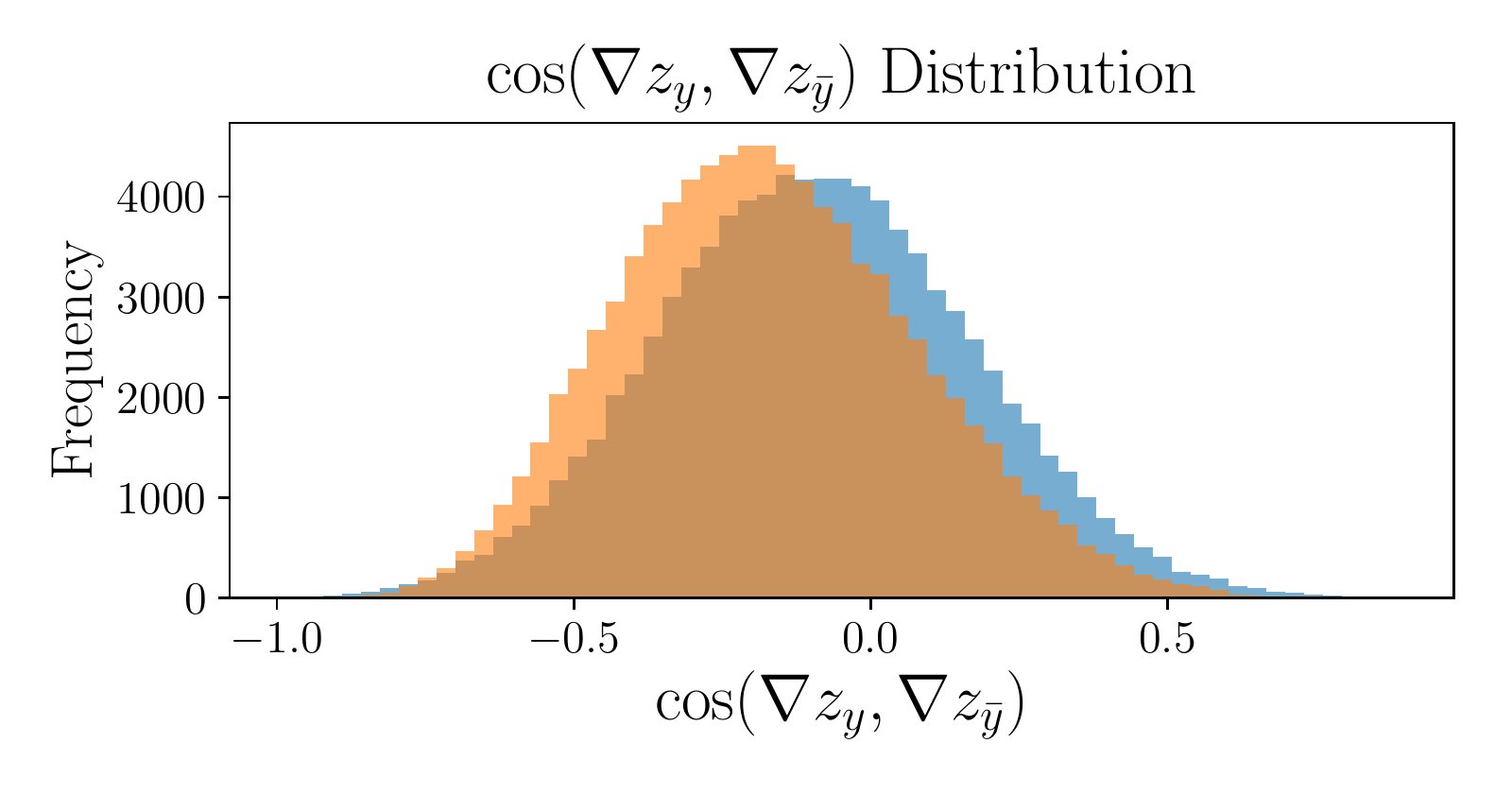}
    \end{subfigure}
    \caption{The effect of adversarial training on the numerator and denominator of the expression for $\epsilon_L$ on CIFAR-10. (left-top) The numerator of Equation~\ref{eq:eps_smallest} (i.e, the logit gap). (left-bottom) The denominator of Equation~\ref{eq:eps_smallest}. (right-top) A histogram of values of $\epsilon_L$ calculated by applying Equation~\ref{eq:eps_smallest} to test data. (right-bottom) Cosine between the gradient vectors of the logits (i.e., gradient coherence).}
    \label{fig:c10_eps_natNadv}
\end{figure}

Similarly to what we observed about adversarial training on MNIST, adversarial training on CIFAR-10 works by greatly shrinking the adversarial gradients and also shrinking the logit gaps. The shrink in the gradients is much more dramatic than the shrink in the logit gap, and overwhelms the decrease in the numerator of Equation~\ref{eq:eps_smallest}. See  \cref{fig:c10_eps_natNadv}.

\section{The magic of Gaussian augmentation plus logit squeezing}\label{appendix:logit_squeeze}

\begin{figure}[!htbp]
    \centering
    \begin{subfigure}[b]{0.49\textwidth}
        \includegraphics[width=.9\textwidth]{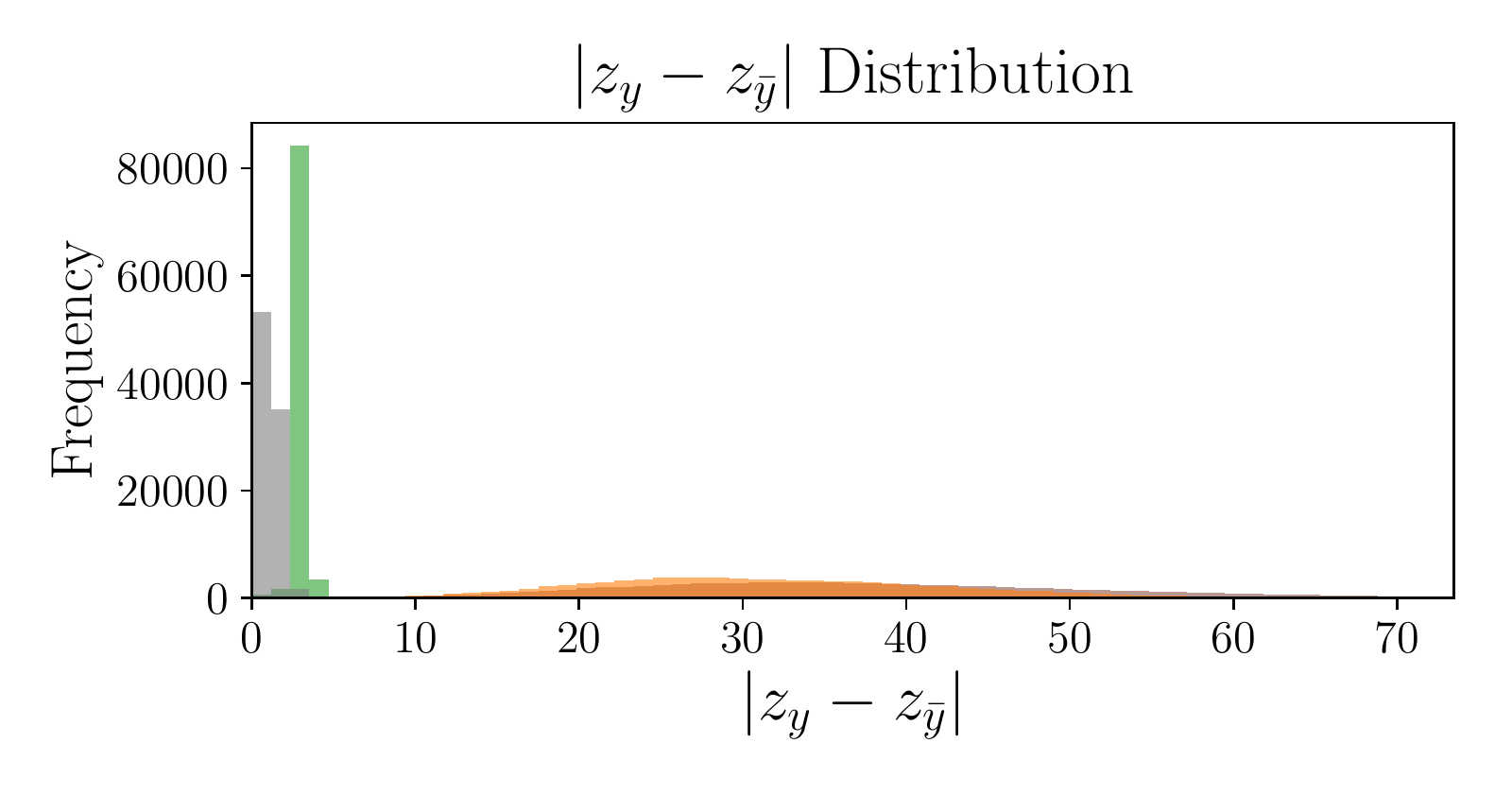}
    \end{subfigure}
    \begin{subfigure}[b]{0.49\textwidth}
        \includegraphics[width=.9\textwidth]{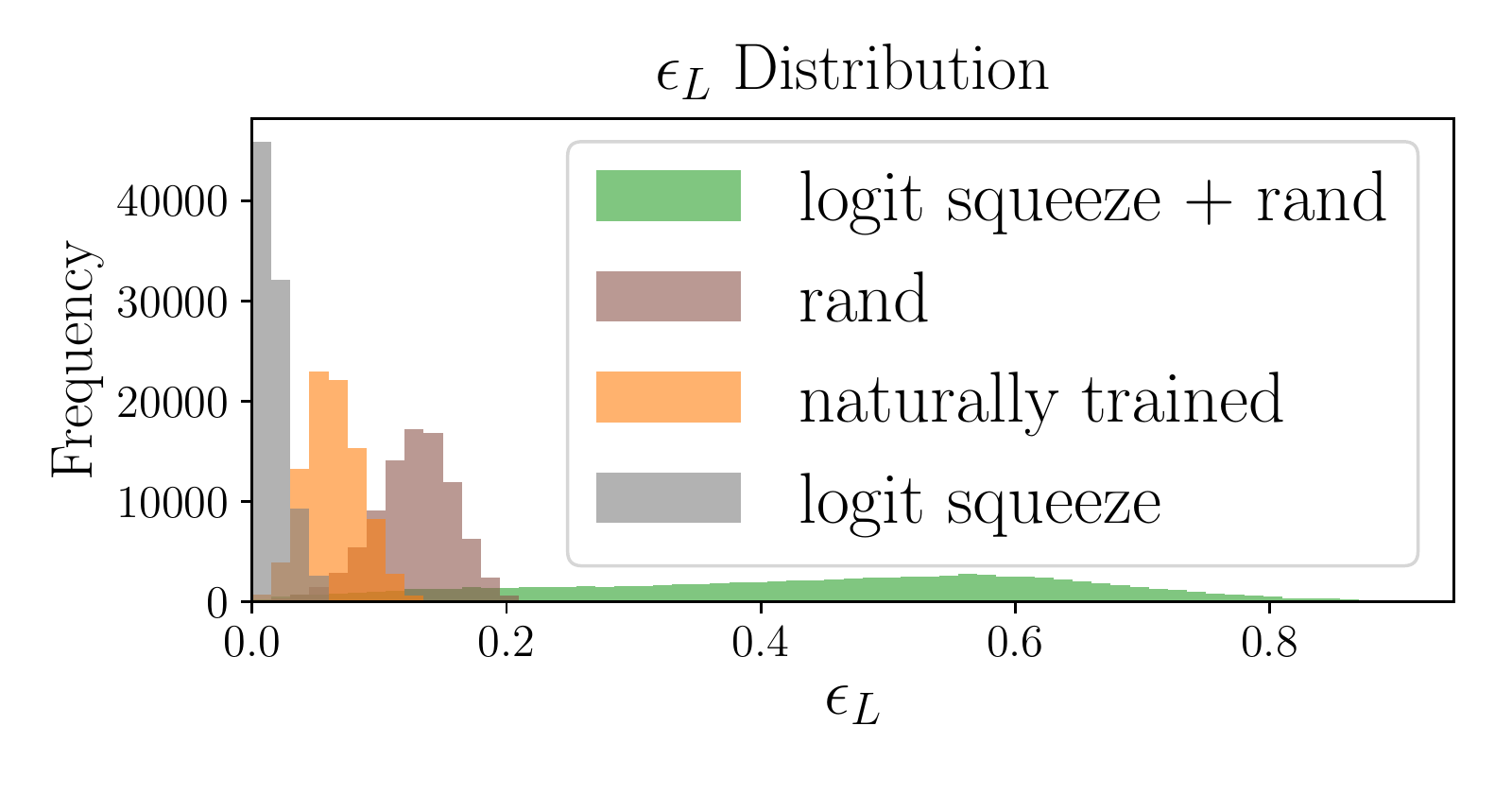}
    \end{subfigure}
        \centering
    \begin{subfigure}[b]{0.49\textwidth}
        \includegraphics[width=.9\textwidth]{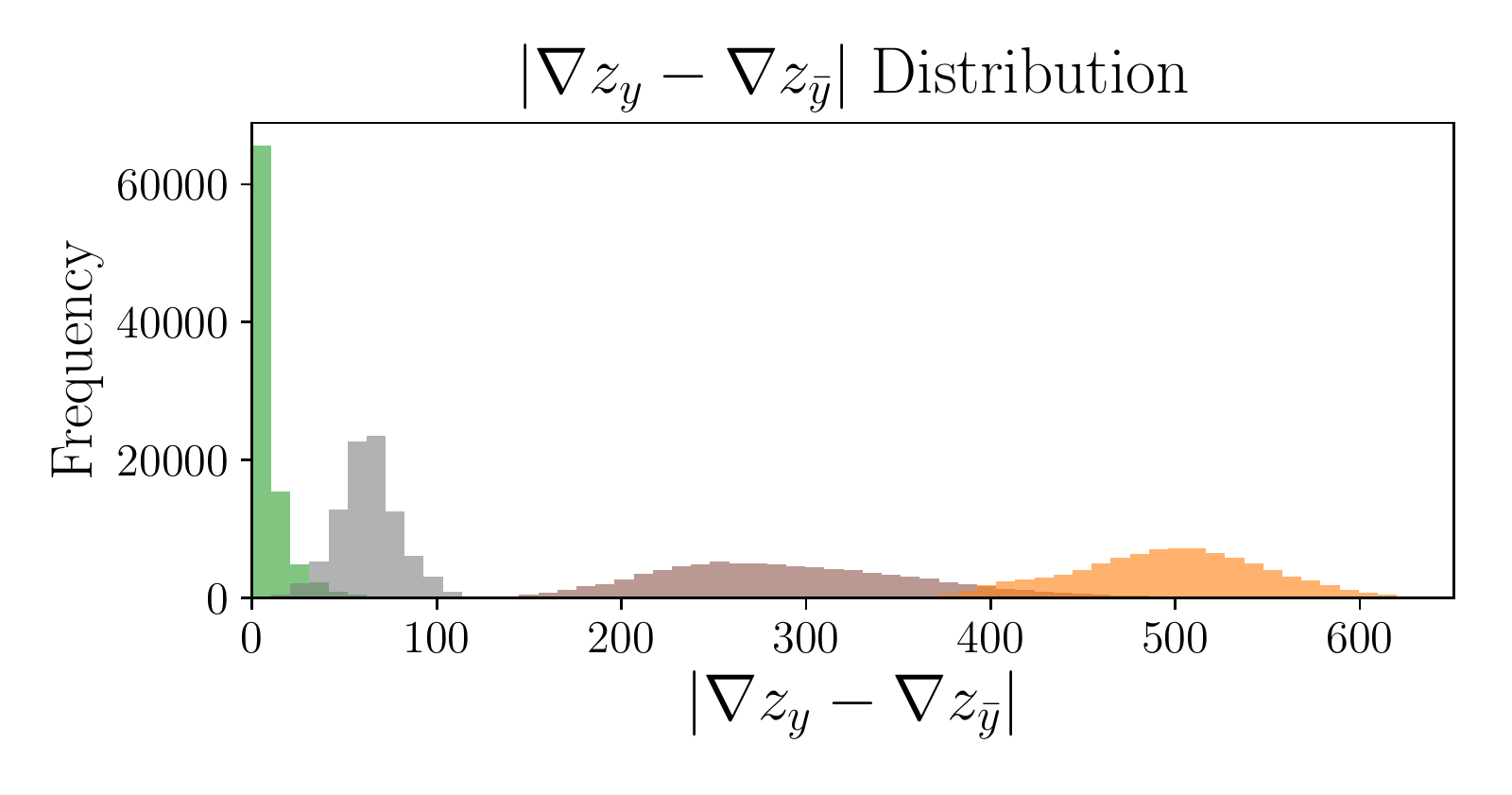}
    \end{subfigure}
    \begin{subfigure}[b]{0.49\textwidth}
        \includegraphics[width=.9\textwidth]{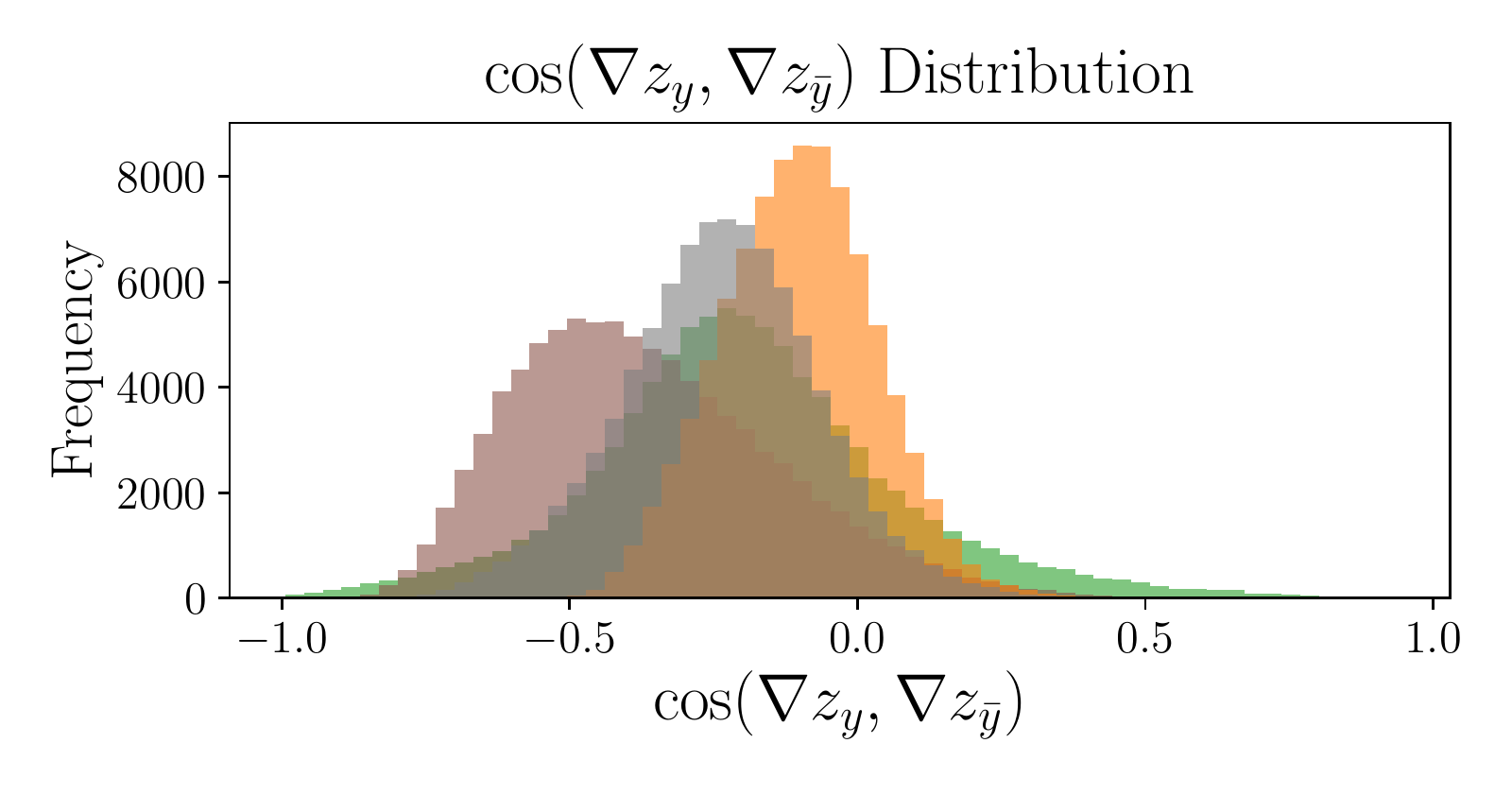}
    \end{subfigure}
    \caption{Random Gaussian augmentation helps logit squeezing. Graphs are for MNIST test examples with $\sigma=0.5$ and $\beta=0.5$. }
    \label{fig:mnist_logit_squeeze_vs_rand}
\end{figure}

As shown empirically in \cref{tab:mnist_logit_squeeze_and_label_smooth_fullV}, and analytically using the linear approximation in Equation~\ref{eq:eps_smallest} evaluated in \cref{fig:mnist_logit_squeeze_vs_rand}, logit squeezing worsens robustness when Gaussian augmentation is not used. However, when fused with Gaussian augmentation, logit squeezing achieves good levels of robustness. This addition of Gaussian augmentation has three observed effects: the gradients get squashed, the logit gap increases, and the gradients get slightly more aligned. The increase in the logit gaps increases the numerator in Equation~\ref{eq:eps_smallest}. This gives a slight edge to logit squeezing in comparison to label smoothing, that mostly works by decreasing the denominator in Equation~\ref{eq:eps_smallest}.

\begin{figure}[!htpb]
    \centering
    \includegraphics[width=.9\textwidth]{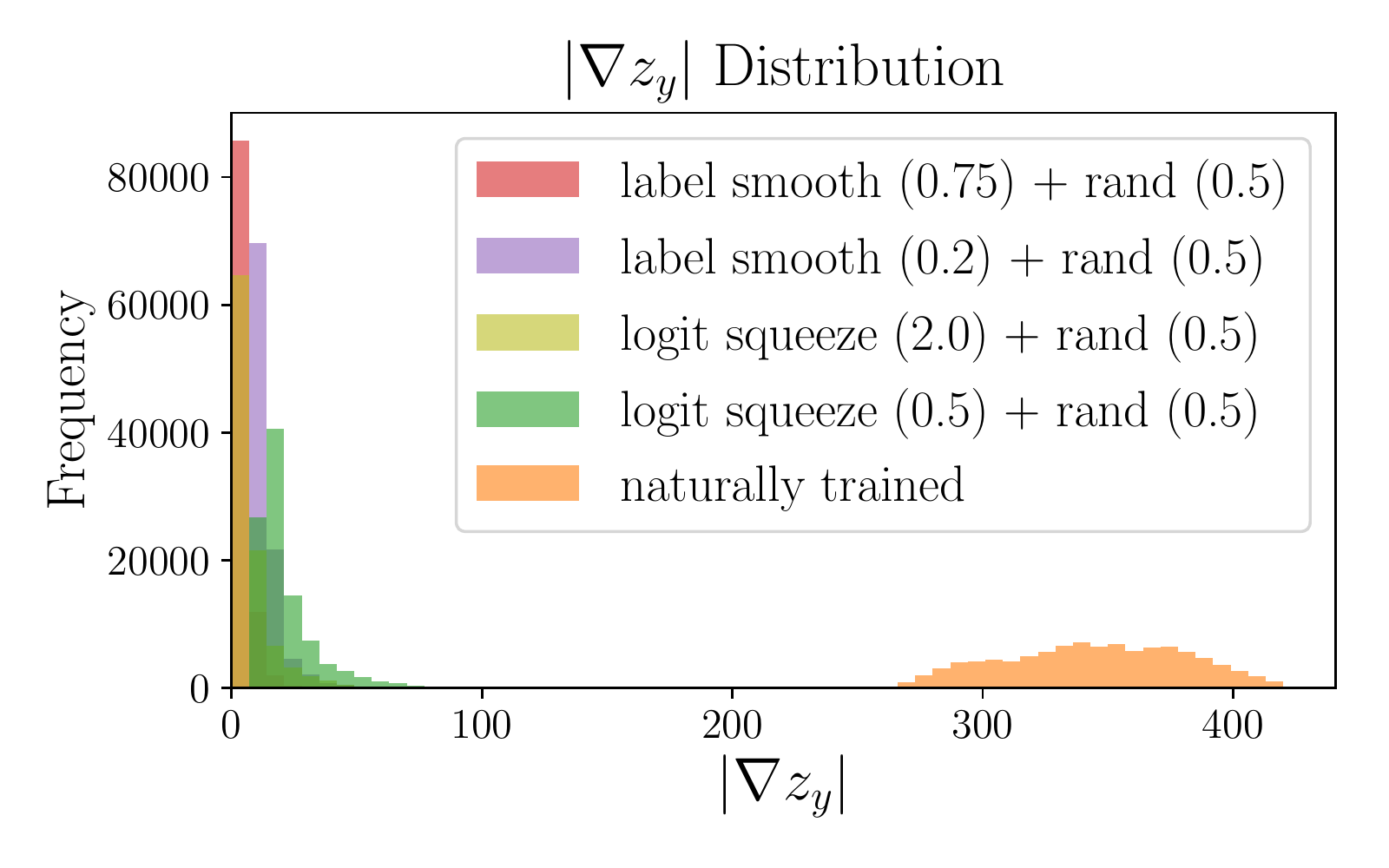}
    \caption{ Label smoothing decreases the difference between the adversarial gradients (the denominator of Equation~\ref{eq:eps_smallest}) by shrinking the gradient magnitudes.  The gradients magnitudes are depicted here using a histogram. }
    \label{fig:1norm_grads_dist}
\end{figure}

\section{Full results on MNIST} \label{appendix:mnist_res}
The results for all of our experiments on MNIST are summarized in \cref{tab:mnist_logit_squeeze_and_label_smooth_fullV}. As can be seen, Gaussian random augmentation by itself ($\beta=\alpha=0$) is effective in increasing robustness on black-box attacks. It does not, however, significantly increase robustness on white-box attacks. Models that are only trained with either logit squeezing or label smoothing without random Gaussian augmentation ($\sigma=0$), can be fooled by an adversary that has knowledge of the model parameters (white-box) with accuracy 100\%. In the black-box setting, they are also not robust. 

Increasing the magnitude of the noise ($\sigma$) generally increases robustness but degrades the performance on the clean examples. Keeping the number of iterations $k$ constant, for extremely large $\sigma$ values, the robustness also starts to drop. At the expense of extra computations $k$, the robustness and the accuracy on clean examples improves.

\begin{table}[!htbp]
    \centering
    \begin{tabular}{c|c|c|c|c|c|c|c|c|c}
    \multicolumn{4}{c}{defense params} & & & \multicolumn{2}{c}{White-box} & \multicolumn{2}{c}{black-box}\\
    & & & & &  &\multicolumn{2}{c}{40-step PGD} &  & \\
    $\alpha$ & $\beta$ & $\sigma$ & $k$ & Train &  Test & X-ent & CW & FGSM & PGD\\
    \hline & & & & & & & & \\
    0 & 0 & 0.5 & 100k & 99.89\% & 98.97\% & 5.18\% & 1.95\% & 84.23\% & 86.21\%\\
    0 & 0 & 0.3 & 400k & 99.97\% & 99.07\% & 3.27\% & 0.83\% & 79.36\% & 81.61\%\\
    0 & 0 & 0.5 & 400k & 99.99\% & 99.09\% & 11.24\% & 17.52\%& 87.28\% & 89.44\%\\
    
    0 & 0.5 & 0.5 & 100k & 99.83\% & 99.06\% & 57.57\% & 55.77\%& 85.91\% & 87.48\%\\
    0 & 0.5 & 0.5 & 400k & 99.96\% & 99.21\% & 72.44\% & 72.39\%& 87.58\% & 89.48\%\\
    
    0 & 2.0 & 0.0 & 400k & 63.11\% & 64.14\% & 0.00\% & 0.00\% & 17.69\% & 24.52\%\\
    0 & 2.0 & 0.3 & 400k & 99.99\% & 97.17\% & 76.08\% & 76.28\% & 82.11\% & 85.80\%\\
    0 & 2.0 & 0.5 & 100k & 99.69\% & 99.05\% & 67.93\% & 66.38\% & 85.09\% & 86.29\%\\
    0 & 2.0 & 0.5 & 400k & 99.94\% & 99.21\% & 79.13\% & 78.36\% & 87.75\% & 89.47\%\\
    
    0 & 5.0 & 0.5 & 100k & 99.27\% & 99.02\% & 65.53\% & 62.00\% & 85.53\% & 86.64\%\\
    0 & 5.0 & 0.5 & 400k & 99.84\% & 99.18\% & 78.21\% & 77.29\% & 87.73\% & 89.25\%\\
    
    0 & 10.0 & 0.5 & 100k & 74.16\% & 75.76\% & 3.15\% & 2.43\% & 57.35\% & 61.55\%\\
    0 & 10.0 & 0.5 & 400k & 13.27\% & 12.93\% & 9.40\% & 8.88\% & 10.60\% & 11.00\%\\
    
    \hline
    
    0 & 0 & 0.3 & 400k & 100.00\% & 99.42\% & 2.90\% & 1.06\% & 78.67\% & 80.67\%\\
    0 & 0 & 0.5 & 400k & 100.00\% & 99.23\% & 11.75\% & 19.64\% & 87.57\% & 89.71\%\\
    0 & 0 & 0.7 & 400k & 99.90\% & 99.08\% & 13.68\% & 22.95\% & 87.53\% & 89.14\%\\
    0 & 0 &1.0 & 400k & 98.66\% & 98.17\% & 1.85\% & 9.33\% & 82.19\% & 83.09\%\\
    
    0.2 & 0 & 0 & 400k & 100.00\% & 99.36\% & 0.00\% & 0.00\% & 23.88\% & 29.33\%\\
    0.2 & 0 & 0.1 & 400k & 100.00\% & 99.39\% & 15.59\% & 17.91\% & 59.14\% & 63.73\%\\
    0.2 & 0 & 0.2 & 400k & 100.00\% & 99.41\% & 55.50\% & 56.63\% & 69.40\% & 76.46\%\\
    0.2 & 0 & 0.3 & 400k & 100.00\% & 99.40\% & 70.97\% & 73.71\% & 81.60\% & 85.38\%\\
    0.2 & 0 & 0.4 & 400k & 100.00\% & 99.33\% & 76.86\% & 80.07\% & 86.71\% & 88.75\%\\
    0.2 & 0 & 0.5 & 400k & 99.99\% & 99.20\% & 76.81\% & 79.55\% & 87.79\% & 89.50\%\\
    0.2 & 0 & 0.6 & 400k & 99.92\% & 99.16\% & 74.28\% & 75.19\% & 87.87\% & 89.43\%\\
    0.2 & 0 & 0.7 & 400k & 99.76\% & 99.04\% & 66.34\% & 70.96\% & 87.27\% & 88.60\%\\
    
    0.5 & 0 & 0.3 & 400k & 100.00\% & 99.36\% & 74.61\% & 75.10\% & 81.66\% & 86.26\%\\
    0.5 & 0 & 0.5 & 400k & 99.97\% & 99.29\% & 79.12\% & 79.24\% & 87.32\% & 89.28\%\\
    0.5 & 0 & 0.7 & 400k & 99.63\% & 98.90\% & 70.17\% & 70.16\% & 86.85\% & 87.49\%\\

    0.75 & 0 & 0 & 400k & 100.00\% & 99.37\% & 0.00\% & 0.00\% & 15.85\% & 21.26\%\\
    0.75 & 0 & 0.3 & 400k & 100.00\% & 99.41\% & 75.21\% & 75.62\% & 79.91\% & 84.10\%\\
    0.75 & 0 & 0.5 & 400k & 99.90\% & 99.23\% & 78.35\% & 79.60\% & 87.20\% & 88.83\%\\
    0.75 & 0 & 0.7 & 400k & 99.44\% & 98.71\% & 70.11\% & 69.95\% & 86.97\% & 87.99\%\\
    
    0.95 & 0 & 0 & 400k & 100.00\% & 99.21\% & 0.00\% & 0.00\% & 16.76\% & 19.81\%\\
    0.95 & 0 & 0.3 & 100k & 99.78\% & 99.38\% & 60.70\% & 61.93\% & 94.29\% & 77.21\%\\
    0.95 & 0 & 0.3 & 400k & 99.99\% & 99.44\% & 74.02\% & 75.46\% & 93.97\% & 85.00\%\\
    0.95 & 0 & 0.3 & 2M & 100.00\% & 99.25\% & 83.60\% & 85.39\% & 85.39\% & 87.22\%\\
    0.95 & 0 & 0.5 & 100k & 99.20\% & 99.01\% & 61.95\% & 60.80\% & 84.83\% & 86.02\%\\
    0.95 & 0 & 0.5 & 400k & 99.82\% & 99.23\% & 74.15\% & 73.19\% & 86.72\% & 87.87\%\\
    0.95 & 0 & 0.5 & 2M & 99.93\% & 98.98\% & 77.92\% & 82.05\% & 87.33\% & 88.78\%\\
    
    \hline
    
    \multicolumn{4}{c}{\cite{madry2017towards} adv tr.} & 100\% & 98.8\% & 93.20\% & 93.9 \% & 96.08\% & 96.81\%\\ 
    \end{tabular}
    \caption{Accuracy of different models trained on MNIST with a 40 step PGD attack on the cross-entropy (X-ent) loss and the Carlini-Wagner (CW) loss under the white-box and black-box threat models. Attacks are $\ell_\infty$ attacks with a maximum perturbation of $\epsilon=0.3$. The iterative white-box attacks have an initial random step. The naturally trained model was used for generating the attacks for the black-box threat model. We use the CW loss for the FGSM attack in the blackbox case. $k$ is the number of training iterations.}
    \label{tab:mnist_logit_squeeze_and_label_smooth_fullV}
\end{table}

\section{Complete results on CIFAR-10}\label{appendix:cifar_res}
Here we take a deeper look at reguarlized training results for CIFAR-10. The conclusions that can be drawn in this case are parallel with those of MNIST discussed in \cref{appendix:mnist_res}. It is worth noting that while the results of logit squeezing outperform those from label smoothing in the white-box setting, training with large squeezing coefficient $\beta$ often fails and results in low accuracy on test data. This breakdown of training rarely happens for label smoothing (even for very large smoothing parameters $\alpha$).
\begin{table}[!htbp]
    \centering
    \begin{tabular}{c|c|c|c|c|c|c|c|c|c}
    \multicolumn{4}{c}{defense params} & \multicolumn{6}{c}{White-box}\\
    & & & & & &\multicolumn{2}{c}{20-step PGD} & \multicolumn{2}{c}{FGSM} \\
    $\alpha$ & $\beta$ & $\sigma$ & k & Train &  Test & xent & CW & xent & CW \\
    \hline & & & & & & & & & \\
    0.2 & 0 & 20 & 80k & 99.88\% & 92.94\% & 31.27\% & 32.16\% & 75.10\% & 69.88\% \\
    0.2 & 0 & 30 & 80k & 99.67\% & 91.08\% & 38.26\% & 39.26\% & 71.93\% &  67.30\% \\
    0.2 & 0 & 40 & 80k & 98.99\% & 87.97\% & 37.12\% & 36.28\% & 67.22\% & 62.01\%  \\

    0.5 & 0 & 20 & 80k & 99.85\% & 92.57\% & 31.37\% & 32.55\% & 75.76\% & 73.45\%  \\
    0.5 & 0 & 30 & 80k & 99.50\% & 90.82\% & 38.30\% & 39.46\% & 72.83\% & 71.32\%  \\

    0.9 & 0 & 30 & 80k & 99.30\% & 90.59\% & 33.89\% & 33.21\% & 2.04\% & 1.11\%  \\
    
    0.2 & 0 & 30 & 160k & 99.72\% & 90.64\% & 37.67\% & 39.60\% & 76.07\% & 69.48\%  \\

    0.5 & 0 & 30 & 160k & 99.70\% & 90.58\% & 40.22\% & 39.96\% & 74.50\% & 71.55\% \\

    0.8 & 0 & 30 & 160k & 99.73\% & 90.83\% & 39.20\% & 37.73\% & 74.29\% &  72.81\% \\
    
    0.95 & 0 & 20 & 160k & 99.90\% & 92.88\% & 43.00\% & 41.29\% & 75.25\% & 74.51\%  \\

    0.95 & 0 & 30 & 160k & 99.64\% & 90.70\% & 53.93\% & 40.68\% & 64.77\% & 70.38\% \\
    0.95 & 0 & 40 & 160k & 98.94\% & 87.81\% & 49.23\% & 35.93\%  & 61.42\% & 64.95\% \\
    
    0.95 & 0 & 30 & 240k & 99.70\% & 90.55\% & 47.27\% & 37.25\% & 64.36\% & 69.44\%  \\
    
    0.8 & 0 & 30 & 320k & 99.72\% & 90.23\% & 43.51\% & 42.96\% & 74.68\% & 72.66\%  \\
    
    0.95 & 0 & 40 & 320k & 94.10\% & 86.12\% & 46.82\% & 19.27\% & 55.11\% & 45.48\%  \\
    \hline
    0 & 10 & 20 & 80k & 99.77\% & 92.16\% & 45.46\% & 41.39\% & 72.28\% & 71.42\% \\ 
    0 & 10 & 20 & 160k & 99.92\% & 92.68\% & 52.55\% & 48.78\% & 76.37\% & 75.79\% \\ 
    0 & 10 & 30 & 80k & 99.45\% & 89.89\% & 48.46\% & 45.51\% & 68.19\% & 67.25\% \\ 
    0 & 10 & 30 & 160k & 99.82\% & 90.49\% & 52.30\% & 49.73\% & 72.08\% & 71.08\% \\ 
    \hline
    \multicolumn{4}{c}{\cite{madry2017towards}} & 100.00\% & 87.25\% & 45.84\% & 46.90\% & 56.22\% & 55.57\% \\

    \end{tabular}
    \caption{White-box attacks on the CIFAR-10 models. All attacks are $\ell_{\infty}$ attacks with $\epsilon=8$. For the 20-step PGD, similar to \cite{madry2017towards}, we use an initial random perturbation.
    }
    \label{tab:c10_white}
\end{table}

\begin{table}[!htbp]
    \centering
    \begin{tabular}{c|c|c|c|c|c|c|c|c|c}
    \multicolumn{4}{c}{defense params} & \multicolumn{6}{c}{Black-box}\\
    & & & & \multicolumn{2}{c}{7-step PGD} &\multicolumn{2}{c}{7-step PGD+Rand} & \multicolumn{2}{c}{FGSM} \\
    $\alpha$ & $\beta$ & $\sigma$ & k & xent &  CW & xent & CW & xent & CW \\
    \hline & & & & & & & & & \\
    0.2 & 0 & 20 & 80k & 71.74\% & 72.44\% & 72.89\% & 73.73\% & 74.51\% & 74.99\% \\
    0.2 & 0 & 30 & 80k & 67.53\% & 68.26\% & 68.46\% & 69.32\% & 70.51\% &  71.36\% \\
    0.2 & 0 & 40 & 80k & 64.39\% & 65.42\% & 65.34\% & 62.01\% & 67.13\% & 68.01\%  \\

    0.5 & 0 & 20 & 80k & 71.17\% & 71.88\% & 72.45\% & 72.93\% & 74.00\% & 74.65\%  \\
    0.5 & 0 & 30 & 80k & 67.31\% & 68.25\% & 68.36\% & 69.44\% & 70.32\% & 70.96\%  \\

    0.9 & 0 & 30 & 80k & 10.07\% & 10.08\% & 10.08\% & 10.02\% & 10.07\% & 10.04\%  \\
    
    0.2 & 0 & 30 & 160k & 67.83\% & 68.72\% & 68.73\% & 69.48\% & 70.66\% & 71.43\%  \\

    0.5 & 0 & 30 & 160k & 67.91\% & 68.74\% & 68.77\% & 69.89\% & 70.51\% & 71.37\% \\

    0.8 & 0 & 30 & 160k & 67.85\% & 68.86\% & 68.88\% & 69.69\% & 70.51\% &  71.29\% \\
    
    0.95 & 0 & 20 & 160k & 71.58\% & 71.96\% & 72.44\% & 73.16\% & 74.01\% & 74.68\%  \\

    0.95 & 0 & 30 & 160k & 68.33\% & 68.88\% & 69.09\% & 69.63\% & 70.85\% & 71.71\% \\
    0.95 & 0 & 40 & 160k & 63.99\% & 64.80\% & 65.13\% & 65.83\%  & 66.98\% & 67.91\% \\
    
    0.95 & 0 & 30 & 240k & 67.88\% & 68.59\% & 68.84\% & 69.63\% & 70.53\% & 71.32\%  \\
    
    0.8 & 0 & 30 & 320k & 67.55\% & 68.32\% & 68.49\% & 69.28\% & 70.03\% & 70.89\%  \\
    
    0.95 & 0 & 40 & 320k & 64.20\% & 65.01\% & 64.90\% & 65.91\% & 66.83\% & 67.66\%  \\
    \hline
    0 & 10 & 20 & 80k & 70.27\% & 70.78\% & 71.29\% & 71.86\% & 72.47\% & 73.30\% \\ 
    0 & 10 & 20 & 160k & 70.75\% & 71.49\% & 71.63\% & 72.22\% & 73.48\% & 73.82\%\\ 
    0 & 10 & 30 & 80k & 66.53\% & 67.46\% & 67.52\% & 68.40\% & 69.30\% & 69.99\% \\ 
    0 & 10 & 30 & 160k & 67.05\% & 67.79\% & 68.30\% & 68.89\% & 69.94\% & 70.61\%\\ 
    \hline
    \multicolumn{4}{c}{\cite{madry2017towards}} & 63.39\%* & 64.38\%* & 63.39\%* & 64.38\%* & 67.00\% & 67.25\% \\

    \end{tabular}
    \caption{Black-box attacks on the CIFAR-10 models. All attacks are $\ell_{\infty}$ attacks with $\epsilon=8$. Similar to \cite{madry2017towards}, We build 7-step PGD attacks and FGSM attacks for the public adversarial trained model of MadryLab. We then use the built attacks for attacking the different models. *: Since we do not have the Madry model, we cannot evaluate it under the PGD attack with and without random initialization and therefore we use the same value that is reported by them for both.}
    \label{tab:c10_black}
\end{table}

\section{Other metrics of similarity between adversarial training, logit squeezing, and label smoothing}
\begin{figure}[!htbp]
    \centering
    \includegraphics[width=.8\textwidth]{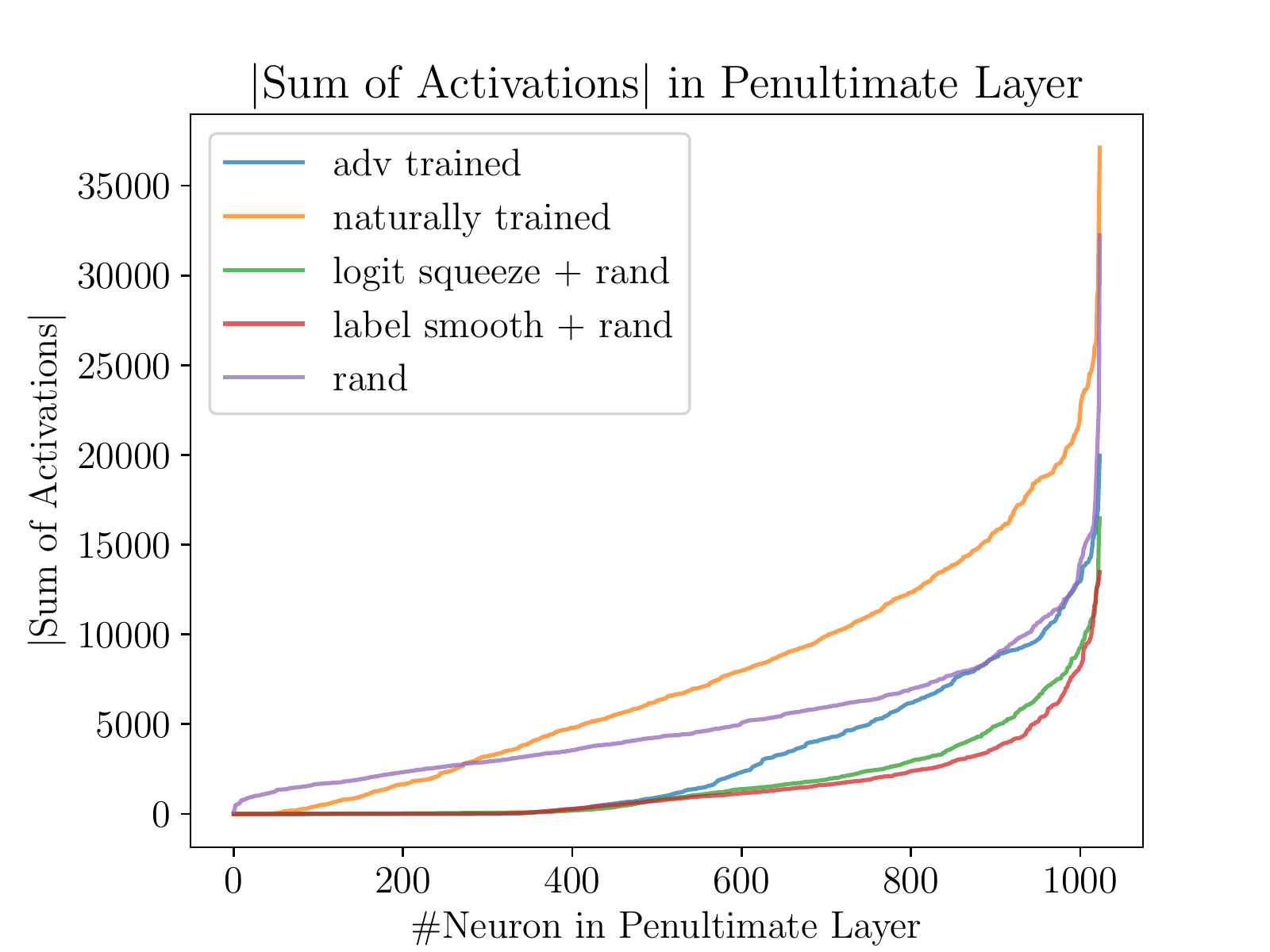}
    \caption{For MNIST, we plot the cumulative magnitude of activations for all neurons in feature layer of a network produced by natural training, adversarial training, natural training with random noise, label smoothing ($\alpha=0.2$) with  random noise, and logit squeezing ($\beta=0.5$) with random noise. In all cases, the noise is Gaussian with $\sigma=0.5$. Interestingly, the combination of Gaussian noise and label smoothing, similar to the combination of Gaussian noise and logit squeezing, deactivates roughly 400 neurons. This is similar to adversarial training. In some sense it seems that all three methods are causing the ``effective'' dimensionality of the deep feature representation layer to shrink.}
    \label{fig:mnist_features_activation}
\end{figure}

\begin{figure}[!htbp]
    \centering
    \includegraphics[width=.8\textwidth]{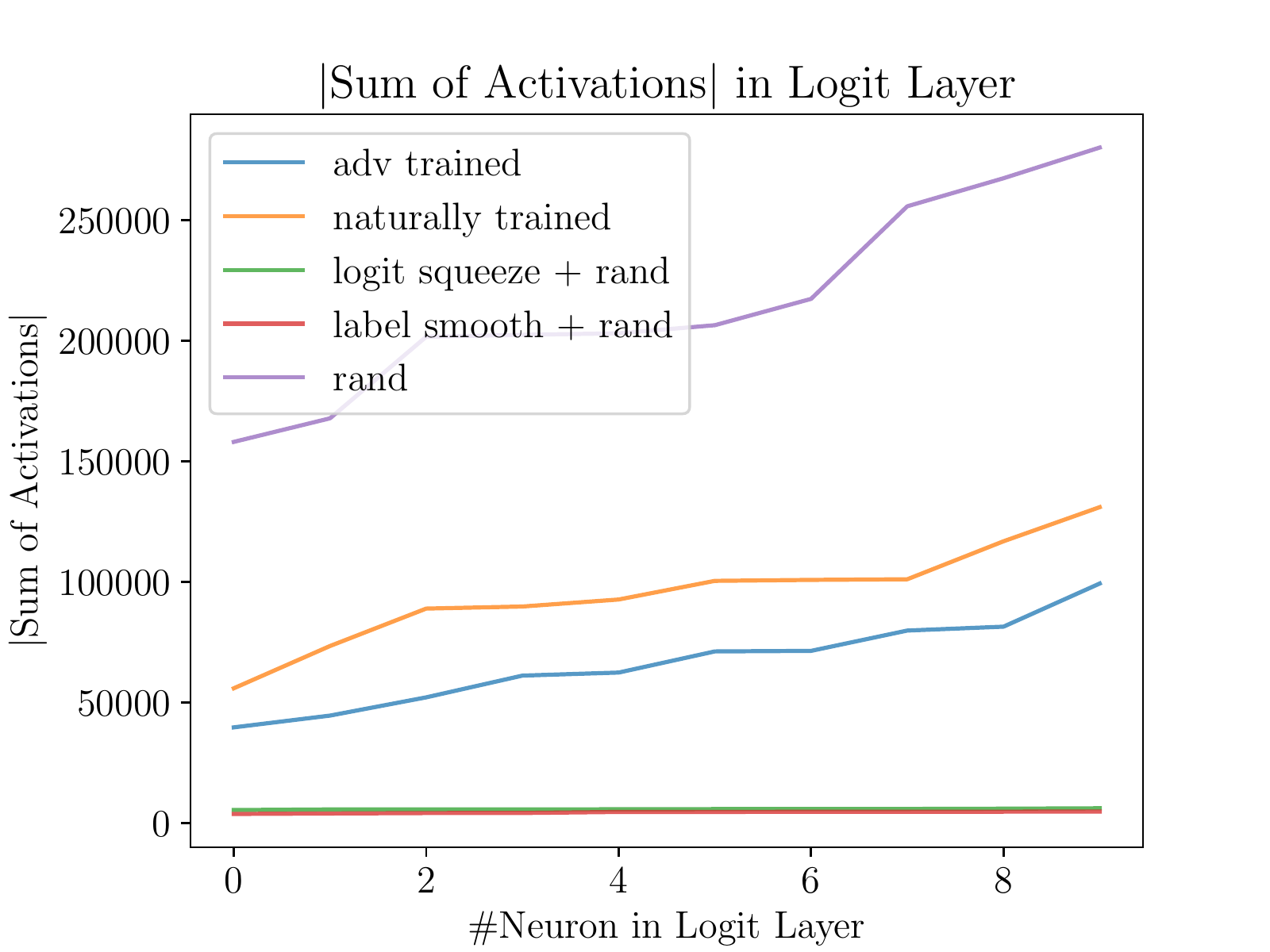}
    \caption{For MNIST, we plot the cumulative sum of activation magnitudes for all neurons in logit layer of a network produced by natural training, adversarial training, natural training with random noise, label smoothing ($LS=0.2$) with  random noise, and logit squeezing ($\beta=0.5$) with random noise. In all cases, the noise is Gaussian with $\sigma=0.5$.}
    \label{fig:mnist_logits_activations}
\end{figure}

While it seems that logit squeezing, label smoothing, and adversarial training have a lot in common when we look at quantities affecting the linear approximation $\epsilon_L$, we wonder whether they still look similar with respect to other metrics. Here, we look at the sum of the activations in the logit layer for every logit (\cref{fig:mnist_logits_activations}) and the sum of activations for every neuron of the penultimate layer (\cref{fig:mnist_features_activation}). The penultimate activations are often seen as the ``feature representation'' that the neural network learns. By summing over the absolute value of the activations of all test examples for every neuron in the penultimate layer, we can identify how many of the neurons are effectively inactive. 

When we perform natural training, all neurons become active for at least some images. After adversarial training, this is no longer the case. Adversarial training is causing the {\em effective} dimensionality of the deep feature representation layer to decrease. One can interpret this as adversarial training learning to ignore features that the adversary can exploit (~400 out of the 1024 neurons of the penultimate layer are deactivated). Shockingly, both label smoothing and logit squeezing do the same -- they deactivate roughly 400 neurons from the deep feature representation layer.

\section{Is our logit-squeezed model giving a false sense of security by breaking the PGD attack?}
As a sanity check, and to verify that the robustness of our models are not due to degrading the functionality of PGD attacks, here we verify that our models indeed have zero accuracy when the adversary is allowed to make huge perturbations. In \cref{tab:unbounded-epsilon-effect} by performing an unbounded PGD attack on a sample logit-squeezed model for the CIFAR-10 ($\beta=10$, $\sigma=30$, and $k=160$k), we verify that our attack is not benefiting from obfuscating the gradients. 

\begin{table}[!htpb]
    \centering
    \begin{tabular}{c|c|c|c}
         PGD steps &  step-size & $\epsilon$ & accuracy \\
         \hline
         30 & 2 & 50 & 14.48\% \\ 
         40 & 2 & 60 & 9.02\% \\ 
         50 & 3 & 100 & 2.07\% \\ 
         50 & 4 & 150 & 0.22\% \\ 
         100 & 4 & 150 & 0.13\% \\ 
         200 & 2 & 255 & 0.00\% \\ 
    \end{tabular}
    \caption{The effect of unbounded $\epsilon$ on the accuracy. The decline in the accuracy as a sanity check shows that the sample model is at least not completely breaking the PGD attack and is not obfuscating the gradients. }
    \label{tab:unbounded-epsilon-effect}
\end{table}

\section{Stronger adversaries: multiple random restarts} \label{appendix:restarts}
We attack our sample hardened CIFAR-10 model ($\beta=10$, Gaussian augmentation $\sigma=30$ model, and $k=160$k iterations), by performing many random restarts. Random restarts can potentially increase the strength of the PGD attack that has a random initialization. As shown in \cref{tab:num_restarts}, increasing the number of random restarts does not significantly degrade the robustness. It should be noted that attacking with more iterations and more random restarts hurts adversarially trained models as well (see the leaderboard in \href{https://github.com/MadryLab/cifar10_challenge}{Madry's Cifar10 Github Repository}). 
\begin{table}[!htbp]
    \centering
    \begin{tabular}{c|c}
         No of Random Restarts & Accuracy \\
         \hline
         1 & 52.3\% \\ 
         2 & 50.84\% \\ 
         3 & 50.43\% \\ 
         4 & 50.24\% \\ 
         5 & 50.21\% \\ 
         6 & 50.13\% \\ 
         7 & 50.04\% \\ 
         8 & 49.86\% \\ 
         9 & 49.86\% \\ 
    \end{tabular}
    \caption{The effect of the number of random restarts while generating the adversarial examples on the accuracy of a model trained with the logit squeezing. It shows that the accuracy plateaus at 9 random restarts.}
    \label{tab:num_restarts}
\end{table}

\section{Stronger adversaries: increasing the number of PGD steps}
Another way to increase the strength of an adversary is by increasing the number of PGD steps for the attack. We notice that increasing the number of steps does not greatly affect the robustness of our sample logit-squeezed model (See \cref{tab:pgd_iteration_accuracy}).  

\begin{table}[th]
    \centering
    \begin{tabular}{c|c}
    Number of PGD steps & Accuracy\\
    \hline
    20 & 49.73\% \\
    40 & 45.33\% \\ 
    100 & 42.36\% \\ 
    400 & 41.58\% \\ 
    1000 & 40.94\% \\ 
    \end{tabular}
    \caption{The effect of the number of steps of the white-box PGD attack on the CW loss (worst case based on \cref{tab:c10_white_Vabs}) for the model trained in $160k$ steps with logit squeezing parameters $\beta=10$ and $\sigma=30$ on CIFAR-10 dataset. The model remains resistant against $\ell_\infty$ attacks with $\epsilon=8$ }
    \label{tab:pgd_iteration_accuracy}
\end{table}
\setcounter{table}{2}

\section{The loss landscape of the logit-squeezed model by moving in an adversarial direction} \label{appendix:loss_land}
\begin{figure}
    \centering
    \includegraphics[width=\textwidth]{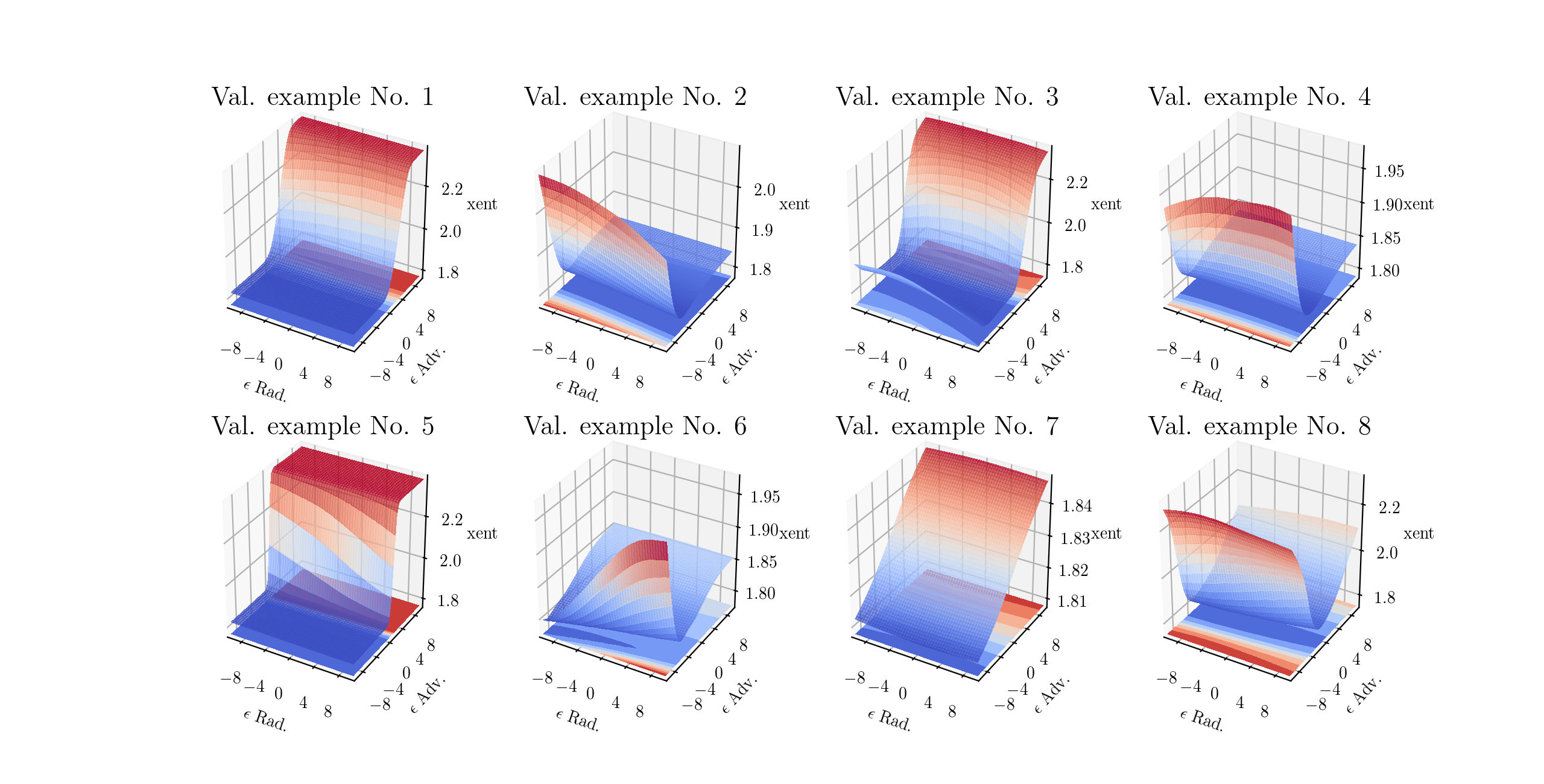}
    \caption{The cross-entropy landscape of the first eight images of the validation set for the model trained for 160k iteration and hyper-parameters $\beta=10$ and $\sigma=30$.
    To plot the loss landscape we take walks in one random direction $r_1$ $\sim$ 
    Rademacher(0.5) and the adversarial direction $a_2 = sign (\nabla_x xent)$ where $xent$ is the cross-entropy loss. We plot the cross-entropy (\textit{i.e. xent}) loss at different points $x = x_i + \epsilon_1 \cdot r_1 + \epsilon_2 \cdot a_2$. Where $x_i$ is the clean image and $  -10\leq \epsilon_1, \epsilon_2 \leq 10$.
    As it can be seen moving along the adversarial direction changes the loss value a lot and moving along the random direction does not make any significant major changes.}
    \label{fig:rad_adv_directions}
\end{figure}

Similar to \cref{fig:rad_drections}, we plot the classification loss landscape surrounding the input images for the first eight images of the validation set. Unlike in \cref{fig:rad_drections} which we changed the clean image along two random directions, in \cref{fig:rad_adv_directions}, we wander around the clean image by moving in the adversarial direction and one random direction. From \cref{fig:rad_adv_directions} we observe that the true direction that the classification loss changes is along the adversarial direction which illustrates that the logit-squeezed model is not masking the gradients.

\end{document}